\documentclass[12pt]{article}

\usepackage{newtxtext,newtxmath}
\RequirePackage[colorlinks=true, allcolors=black]{hyperref}

\usepackage{graphicx}

\usepackage[letterpaper,margin=1in]{geometry}

\renewenvironment{abstract}
	{\quotation}
	{\endquotation}

\date{}

\makeatletter
\renewcommand{\fnum@figure}{\textbf{Figure \thefigure}}
\renewcommand{\fnum@table}{\textbf{Table \thetable}}
\makeatother

\usepackage{scicite}

\usepackage{url}

\newcommand{\bm}[1]{\boldsymbol{#1}}

\def\scititle{
	Analogical Learning for Cross-Scenario Generalization: Framework and Application to Intelligent Localization
}

\title{\bfseries \boldmath \scititle}

\author{
	Zirui~Chen$^{1,2\dagger}$,
	Hongning~Ruan$^{1,2,3\dagger}$,
	Zhaoyang~Zhang$^{1,2,3\ast}$,
	Ziqing~Xing$^{1,2,3}$,\and
	Ridong~Li$^{1,2}$,
	Zhaohui~Yang$^{1,2}$,
    Mérouane~Debbah$^{4}$\and
	\small$^{1}$College of Information Science and Electronic Engineering, Zhejiang University, Hangzhou 310027, China.\and
	\small$^{2}$Zhejiang Provincial Key Laboratory of Multi-modal Communication Networks\and\small and Intelligent Information Processing, Hangzhou 310027, China.\and
    \small$^{3}$Institute of Fundamental and Transdisciplinary Research, Zhejiang University, Hangzhou 310058, China.\and
    \small$^{4}$Research Institute for Digital Future, Khalifa University, 127788 Abu Dhabi, UAE.\and
	\small$^\ast$Corresponding author. Email: {ning\_ming@zju.edu.cn}\and
	\small$^\dagger$These authors contributed equally to this work.
}

\begin{document} 

\maketitle

\begin{abstract} \bfseries \boldmath
Modern learning systems often struggle with joint learning across diverse scenarios and immediate adaptation to new ones, because they rely heavily on the scenario-dependent absolute data-label representations. 
Here, we propose analogical learning (AL), a learning framework that explores the inherent invariance of the underlying physical processes across scenarios, to improve the cross-scenario generalization. 
Specifically, we introduce the physical concepts of reference frames and relativity into the neural modeling. 
The resultant framework explicitly employs intra-scenario data-label pairs as reference anchors and enforces the network to mediate its data-to-label transformation through data-domain relative metrics that factor out the scenario-dependent variations.
We instantiate AL with Mateformer, a bipartite Transformer-based neural architecture. Each layer of the auxiliary Transformer extracts certain feature space of the current data, while the corresponding layer of the primary Transformer computes attention among the data feature space and then use it as a relativity metric to weight the current label feature to synthesize the next label feature and, ultimately, the final prediction. 
We apply AL to intelligent wireless localization, a representative multi-scenario learning task. Across synthetic, real-world, and city-scale datasets, AL enables robust cross-scenario transfer and multi-scenario joint learning, achieving wavelength-scale localization accuracy that matches or surpasses state-of-the-art methods. This physics-inspired learning framework provides a promising alternative for other cross-scenario learning tasks and applications.
Code is available at \url{https://github.com/ziruichen-research/ALLoc}.
\end{abstract}

\section*{Introduction}

\begin{figure}[!htbp]
	\centering
	\includegraphics[width=\textwidth]{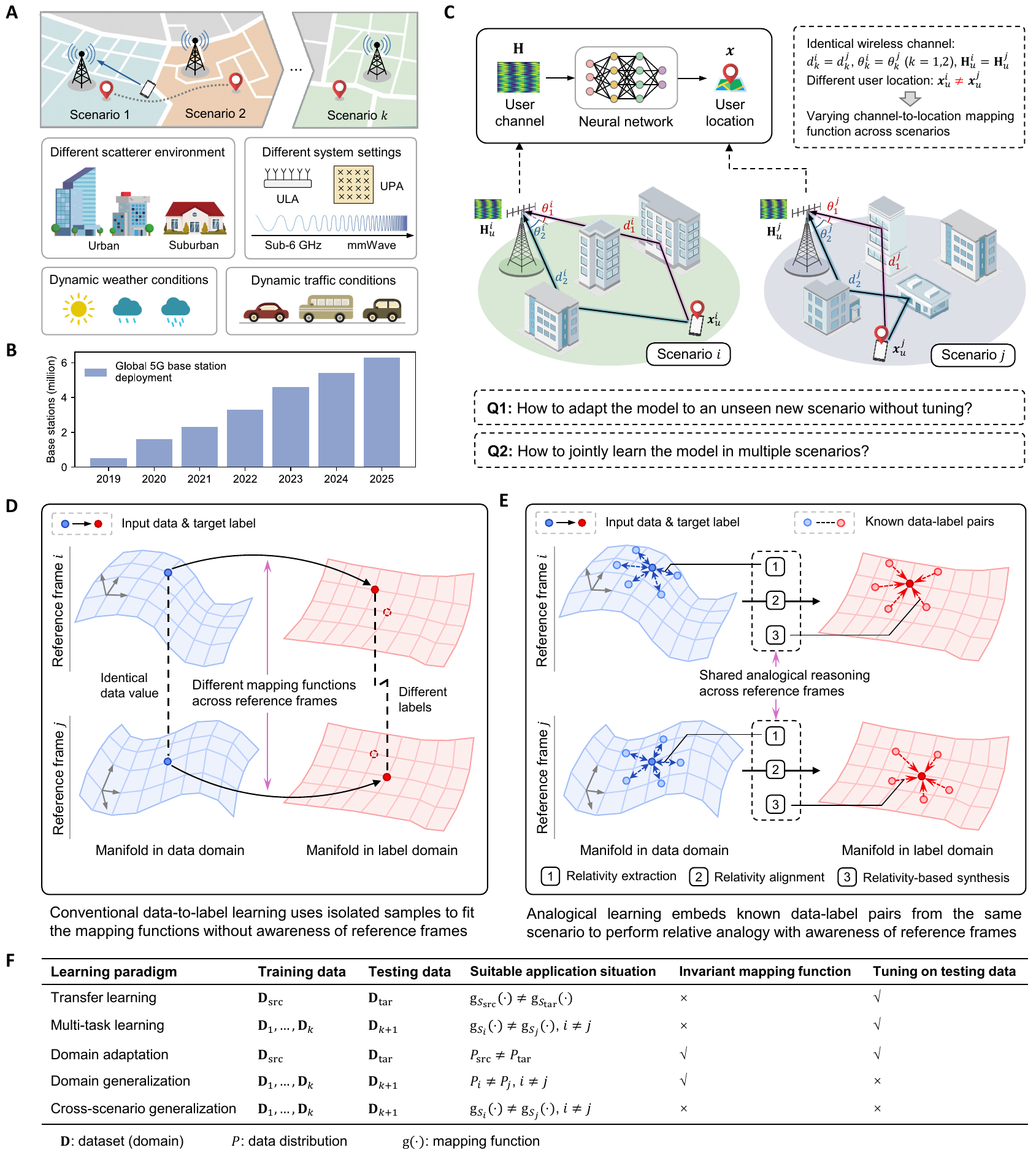}
	\caption{\fontsize{10pt}{9pt}\selectfont \textbf{Backgrounds and motivations of proposed analogical learning.}
		(\textbf{A}) The differences among cellular scenarios, along with dynamic weather and traffic conditions, give rise to diverse electromagnetic characteristics. (\textbf{B}) The number of global 5G base stations. Data retrieved from \cite{5g-base-stations}. (\textbf{C}) Impact of inter-scenario differences on wireless localization: the channel-location mapping function for scenario $i$ conflicts with that for scenario $j$, posing challenges on cross-scenario generalization and multi-scenario learning. (\textbf{D})  Conventional data-to-label learning approaches fit numerical mapping functions, preventing generalization across diverse reference frames. (\textbf{E}) By embedding known data-label pairs within the same scenario and designing effective analogy mechanisms, AL can recognize the reference frames to which the input data belong, addressing existing limitations. (\textbf{F}) Comparison between cross-scenario generalization and related learning paradigms. }
	\label{fig:overview}
\end{figure}

%
Deep learning (DL)-based approaches for scientific and engineering problems often involve cross-scenario generalization, i.e., adapting to a new scenario based on a model learned in other scenarios that share underlying dynamics but differ in spatio-temporal states \cite{dl_book}. In such cases, a typical yet challenging constraint often appears: factors like computational overhead, time-to-deployment latency, and data security prevent the artificial intelligence (AI) model from being fine-tuned in the new scenario, forcing its immediate deployment as a frozen, off-the-shelf solution. For instance, developers of autonomous driving solutions can only train their models using data from certain regions. When deploying to a new city with strict data localization—where data is only accessible by local vehicles—the model must be applied directly without retraining \cite{loc_importance2,autonomous_driving_2}. 
Furthermore, in learning-based robotic control, policies typically rely on low-cost simulated data for training and are subsequently deployed on inference-only edge devices in the real world. This implies that the policy model cannot undergo retraining, yet it must successfully adapt to entirely unseen scenarios \cite{robot_manipulation,robot_manipulation_2}.
Moreover, as a prevailing example, intelligent wireless localization must deal with the millions of scenarios in wireless cellular networks, wherein the complex scattering environment, diverse system settings, and dynamic conditions such as varying weather or traffic patterns yield vastly different electromagnetic characteristics \cite{wBAIM} (Fig. \ref{fig:overview}A, B). Likewise, developers of wireless networks expect models to generalize directly across scenarios rather than undergoing continuous retraining or fine-tuning, thereby significantly reducing the power consumption and latency burdens. In summary, there is a widespread and pressing need for cross-scenario generalization—that is, the ability to deploy a frozen or general model into arbitrary scenarios.
 

However, this technical requirement is inherently challenging to fulfill. For example, intelligent wireless localization generally relies on a localization neural network (LNN) trained to learn the mapping function from user channel state information (CSI) to user location \cite{ad_cnn1,ad_cnn2_long,ad_cnn3_long,cnn4,mccnet}. However, the diversity between scenarios causes the channel-to-location mapping function to exhibit distinct numerical characteristics. As illustrated in Fig. \ref{fig:overview}C, the same channel data often correspond to different locations due to variations in scattering environments. This implies that DL models based on such function fitting face fundamental challenges in generalizing across multiple scenarios effectively. More specifically, this raises two pivotal questions:

First, \textit{how to adapt the model to an unseen new scenario without tuning}? Due to differences in mapping functions, LNNs trained using data from one scenario often exhibit catastrophic performance degradation when directly deployed in new scenarios \cite{transloc1,transloc2,metaloc}. Existing learning paradigms typically employ adaptation strategies and optimization methods like transfer learning \cite{transfer_learning} and meta-learning \cite{meta_learning}, along with parameter-sharing strategies such as model fine-tuning \cite{multi_task}, to capture shareable learning patterns across scenarios. However, all these approaches invariably require additional training hardware investment and latency overhead to generate new models, which makes it difficult to cope with the vast number and dynamic nature of real-world scenarios.

Second, \textit{how to jointly learn the model in multiple scenarios}? Within each single scenario, due to fixed scattering environment and system setting, as well as limited data collection resources, the training data is often constrained in both quantity and diversity, resulting in suboptimal model performance after single-scenario learning. Combining massive amounts of data from multiple scenarios has the potential to improve the model performance, a data scaling approach that has already been validated in existing large language models (LLMs) and other foundational models. However, due to conflicts in the mapping functions between different scenarios, existing learning architectures are unable to collaboratively train a single localization foundation model across multiple scenarios, thereby failing to fully leverage the multi-scenario data.

To answer these two questions, we first adopt the perspective of physical reference to reveal the root causes of the limitations in existing methods.
In fact, although the numerical expressions of mapping functions vary across scenarios, the inherent relationships between data and labels remain unaffected by external scenario factors. For example, since the electromagnetic propagation laws determine a continuous mapping relationship between the channel and location in a given scenario, the localization task can be interpreted as smoothly unfolding the channel manifold in a high-dimensional data space to obtain a flat label manifold that characterizes the user location. However, due to differences in deployment environments or measurement methods, the channel manifold within each scenario undergoes distinct stretching and rotation, resulting in the same data value potentially being mapped to different labels in different scenarios, as shown in Fig. \ref{fig:overview}D. 
In other words, the physical structure of the channel manifold governs the feasibility of flattening. Yet, characterizing this physical essence cannot rely solely on numerical values; it must be integrated with a scenario-dependent reference frame (reflecting this distinct stretching and rotation) to be accurately conveyed. However, traditional data-to-label learning methods use isolated samples to learn absolute numerical relationships between data and labels. Such numerical information is directly influenced by scenario factors, resulting in models that are applicable only to specific scenarios.

Therefore, to eliminate the influence of external scenario factors, we propose analogical learning (AL), whose core idea is illustrated in Fig. \ref{fig:overview}E. The AL approach no longer relies on isolated samples for learning and inference.
Instead, it additionally embeds known data-label pairs from the studied scenario as references. These embedded samples share the same reference frame with the samples to be inferred, thereby endowing the learning and inference processes with the awareness of reference frames. Specifically, AL performs joint representation learning by evaluating relativity in multiple feature spaces.
This process first extracts relative weights between the input data and embedded references within the data domain, then aligns these weights to the label domain, and finally synthesizes the target label based on the embedded labels under the guidance of these weights. To achieve this, we design a bipartite neural network architecture called \textit{Mateformer}, which utilizes a multi-layer multi-head attention mechanism \cite{attention} to efficiently execute the aforementioned three-stage process. In this architecture, all data are interpreted through their relative relationship (normalized attention weights) with other data in the same scenario rather than through scenario-sensitive absolute numerical values, making the learned patterns robust to scenario-specific variations. This design enables AL to learn collaboratively across multiple heterogeneous scenarios and apply the learned knowledge directly to new scenarios based on local data.

We apply AL to intelligent wireless localization, achieving dual advances: first, allowing the trained model to achieve sub-meter level state-of-the-art accuracy in new scenarios without any additional tuning, and second, enabling joint training across multiple scenarios despite significant multi-scenario differences. These advances significantly enhance the model's reliability against complex environmental factors. For example, AL maintains a localization accuracy during heavy rainfall that is nearly on par with its performance under clear weather, and continuously delivers stable performance under changing traffic and road conditions. Furthermore, AL demonstrates exceptional reliability when applied to real-world data and complex realistic urban scenarios. These capabilities not only offer a promising new direction for intelligent wireless localization, but also provide a fresh framework for the intelligent applications in other multi-scenario systems.

\section*{Results}
\subsection*{Cross-scenario generalization in intelligent wireless localization problem}
Intelligent wireless localization is a typical problem in the next-generation wireless technologies. It leverages electromagnetic state information, such as CSI, obtained from communication between the user and the BS to infer the user’s spatial state, which can be formally expressed as $\bm{x}_u={\mathrm{g}}_{S}(\mathbf{H}_u)$, where $\bm{x}_u$ denotes the relative location between the user and the BS, the absolute location of the user can be obtained by combining it with the absolute coordinates of the BS, $\mathbf{H}_u$ represents the user CSI, and $\mathrm{g}_S(\cdot)$ represents the mapping relationship from CSI to location in a specific cellular scenario $S$. Due to the extreme complexity of electromagnetic interactions between wireless signals and unknown scattering environment, modern application-driven wireless localization generally requires the use of DL techniques. In cellular networks, some devices equipped with high-precision localization capabilities can feed their location information to the BS, which together with the corresponding CSI of the device form CSI-location pairs to build a training dataset $\mathbf{D}=\{(\mathbf{H}_1,\bm{x}_1);\dots;(\mathbf{H}_l,\bm{x}_l)\}$, where $l$ is the number of training data. Based on this dataset, an neural network $NN(\cdot)$ can be employed to learn the data-to-label (CSI-location) mapping function $\mathrm{g}_S(\cdot)$. Once $NN(\cdot)$ finishes training, it can be directly applied to new data in this scenario, possessing the cross-sample generalizability, i.e., predicting $\bm{x}_{l+1}$ that corresponds to a given $\mathbf{H}_{l+1}$. This will significantly reduce the reliance of regular user devices on high-precision localization hardware \cite{cooper_loc}.

However, when intelligent wireless localization techniques are applied at the entire wireless system level, rather than within specific cellular scenarios, the problem becomes significantly more challenging. As illustrated in Fig. \ref{fig:overview}A, there are significant differences among cellular scenarios and the scattering environment within a scenario often changes dynamically. This implies that the channel-to-location mapping function $\mathrm{g}_S(\cdot)$ is no longer fixed, but becomes scenario-dependent, where $S\in\{S_1,\dots,S_k\}$ represents diverse scenarios, and $\{1,\dots,k\}$ are the scenario index. In this case, the numerical characteristics of $\mathrm{g}_S(\cdot)$ vary as the scenario $S$ changes, so that the same $\mathbf{H}_u$ corresponds to different $\bm{x}_u$, as shown in Fig. \ref{fig:overview}B. Since there is no universal mapping function applicable to diverse scenarios, data-to-label learning can only be limited to a specific scenario, i.e., learning a $NN_i(\cdot)$ based on the dataset collected in a specific scenario $\mathbf{D}_i=\{(\mathbf{H}_1,\bm{x}_1)^i;\dots;(\mathbf{H}_{l^i},\bm{x}_{l^i})^i\}$, $i\in\{1,\dots,k\}$, where $i$ is the scenario index, $l^i$ is number of training data in scenario $S_i$. Meanwhile, $NN_{i}(\cdot)$, which approximates $\mathrm{g}_{S_i}(\cdot)$, cannot directly substitute $\mathrm{g}_{S_j}(\cdot)$ (where $j\ne i$) and is therefore inapplicable to other scenarios. As new scenarios continue to emerge and existing ones evolve dynamically, wireless systems will have to continuously train and fine-tune their LNNs, resulting in ever-increasing training costs and frequent obsolescence of intelligence. Thus, achieving multi-scenario generalization is a challenging but urgent demand for system-level intelligent wireless localization. 

To address these challenges, we reformulate $\mathrm{g}_{S}(\cdot)$ as $\mathrm{g}^*(S^*,\cdot)$, where $\mathrm{g}^*(\cdot, \cdot)$ resembles the underlying feature transformation process instead of specific numerical function. Coupling $\mathrm{g}^*(\cdot, \cdot)$ with the scenario specificity $S^*$ constitutes the internal mapping function $\mathrm{g}_{S}(\cdot)$ within a scenario. Notably, this is a conceptual simplification for clarity. $S^*$ may not necessarily be represented by certain numerical equivalents, and thus $\mathrm{g}^*(\cdot, \cdot)$ is an abstract concept rather than a rigorous mathematical function. However, based on $\mathrm{g}^*(\cdot, \cdot)$, we can intuitively observe that scenarios and samples essentially resemble two symmetric dimensions of intelligent learning. If neural networks can learn on numerous samples and then directly predict corresponding labels with new data inputs after learning, could they also learn across multiple scenarios and adapt to new scenarios directly upon acquiring relevant information, without additional training? This hypothesis forms the core objective of this article, which can be formally stated as:

{
\noindent
\hspace*{\fill}
\framebox[0.93\columnwidth]{
    \begin{minipage}{\dimexpr0.92\columnwidth-2\fboxsep-2\fboxrule}
        \textbf{Background}: Each scenario $S_i$ possesses its own dataset $\mathbf{D}_i~(i=1,2,\ldots)$. \\
        \textbf{Learning}: Training a $NN(\cdot)$ with limited scenario datasets $\mathbf{D}_1,\ldots,\mathbf{D}_k$. \\
        \textbf{Deployment}: Without new training, the trained $NN(\cdot)$ can still predict $\bm{x}_{l^{k+1}+1}^{k+1}$, which corresponds to $\mathbf{H}_{l^{k+1}+1}^{k+1}$.
    \end{minipage}
}
\hspace*{\fill}
}
\vspace{0.5em}

\noindent The superscripts of $\mathbf{H}$ and $\bm{x}$ denote the scenario indices, while the subscripts represent the sample indices within the studied scenario. In other words, the intelligence simultaneously achieves generalization to both new samples and new scenarios.

It is important to emphasize that cross-scenario generalization differs fundamentally from domain adaptation \cite{domain_adaptation} or domain generalization \cite{domain_generalization}. Domain adaptation and domain generalization address the challenge arising from the difference between the data distribution $P_i$ in the source domain $\mathbf{D}_i$ and $P_j$ in the target domain $\mathbf{D}_j$, which causes performance degradation when a model trained on $P_i$ is deployed to $P_j$. However, the mapping function $\mathrm{g}(\cdot)$ remains invariant across domains. This is different from the cross-scenario generalization problem, where $\mathrm{g}(\cdot)$ varies across scenarios. Furthermore, while existing approaches such as transfer learning \cite{transfer_learning} or multi-task learning \cite{multi_task} can capture shareable learning patterns in mapping functions across different scenarios, they still require additional training when dealing with scenario-specific characteristics. This contrasts with the requirement of cross-scenario generalization, which necessitates deployment in new scenarios without the need for additional training. An overview comparison of these concepts are shown in Fig. \ref{fig:overview}F, with detailed discussions presented in Supplementary Text.

\subsection*{Analogical learning architecture}
\begin{figure}[!htbp]
	\centering
	\includegraphics[width=\textwidth]{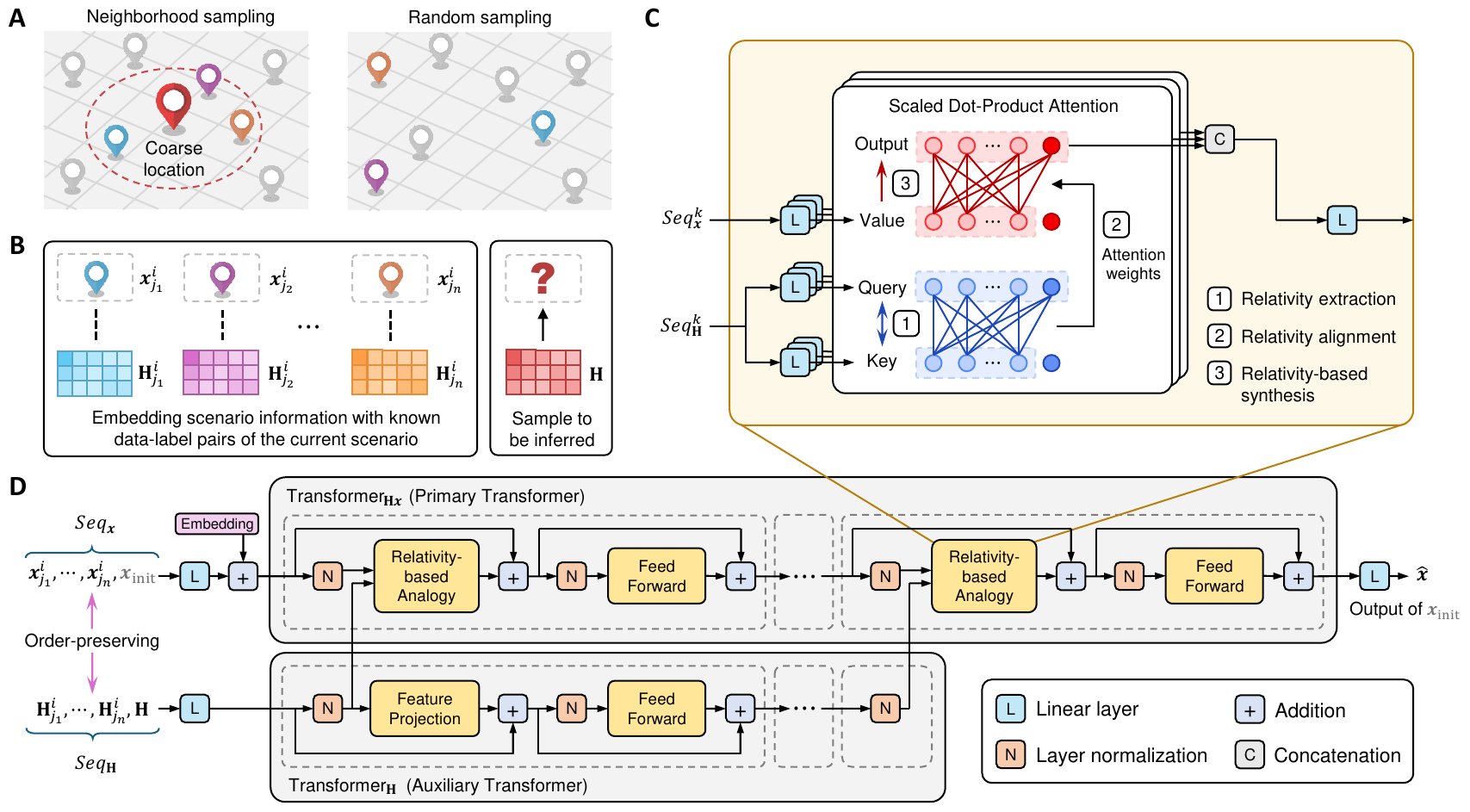}
	\caption{\textbf{The architecture of the proposed analogical learning.}
		(\textbf{A}) Two approaches to select the embedded data-label pairs: neighborhood sampling of the to-be-inferred samples (left) and random sampling from the entire dataset (right). (\textbf{B}) Embedding scenario information as references for analogy. (\textbf{C}) Attention mechanism for extracting relativity weights and guiding the synthesis. (\textbf{D}) The specific network structure of the designed Mateformer.}
	\label{fig:mateformer}
\end{figure}

Inspired by $\mathrm{g}^*(S^*,\cdot)$, we conjectured the feasibility of learning multi-scenario generalization. However, due to the non-characterizability of $S^*$, directly using $NN(S^*,\cdot)$ as a learning objective is infeasible. To address this, we introduce a relaxed approach: instead of taking $S^*$ as input, the neural network can leverage information that is deterministically related to $S^*$, such as data-label pairs from the studied scenario, thereby implicitly capturing the necessary scenario-specific characteristics. Based on these insights, we formulate the following learning objective:
\begin{align}
    &NN((\mathbf{H}_j^i,\bm{x}_j^i)_{j\in\Theta},\mathbf{H})\to\bm{x}, \label{eq:new_paradigm} \\ \nonumber
    &\Theta=\{j_1,\dots,j_n\}\subseteq\{1,\dots,l_i\},
\end{align}
where the user CSI $\mathbf{H}$ belongs to scenario $S_i$. The pair $(\mathbf{H}_j^i,\bm{x}_j^i)$ includes not only the state values, but also the correspondences. In this objective, the reasoning about $\bm{x}$ no longer relies solely on $\mathbf{H}$ but also incorporates the embedded pairs $(\mathbf{H}_j^i,\bm{x}_j^i)$ as references to perceive the characteristics of the scenario, as illustrated in Fig. \ref{fig:mateformer}B. This design enables the trained $NN(\cdot)$ to dynamically adapt to the variability among scenarios.

To limit computational complexity, $\Theta$ is typically not set as the entire collected dataset but rather as an appropriate subset, as shown in Fig. \ref{fig:mateformer}A. In wireless localization task, selecting the neighborhood of the to-be-inferred sample is a reasonable strategy, as neighboring samples typically exhibit higher similarity with the current sample. During training, neighborhood search can be performed using the true value of $\bm{x}$. During inference, since the true value of $\bm{x}$ is unknown, coarse values $\bm{x}'$ provided by global navigation satellite systems (GNSS) \cite{gnss} can be used for pseudo-neighborhood search. Since the pseudo-neighborhood still largely overlaps with the true neighborhood, it can serve as an effective approximation for neighborhood sampling. If neighborhood search is not performed, data-label pairs can still be provided by random sampling from the dataset. The experimental results will demonstrate that localization inference using random sampling can still be effectively performed, albeit with some loss in accuracy. Furthermore, such low-accuracy location can serve as a coarse value $\bm{x}'$ to iteratively initiate new neighborhood sampling. Therefore, while auxiliary information provided by GNSS can provide convenience, it is not indispensable for AL-based wireless localization.

By incorporating known reference anchors into the learning process, the neural network is able to reason based on the relativity between all samples (whether embedded or to be inferred ) rather than their isolated absolute numerical values. Specifically, we perform a transformation to Eq. \ref{eq:new_paradigm}, implicitly representing $\mathbf{H}$ through its relative relationships with $\mathbf{H}_j^i$, and then aligning these relationships to the label space to guide the reconstruction of $\bm{x}$ from $\bm{x}_j^i$. This process can be formulated as follows:
\begin{equation}
    NN_{\rm use\_rela}({NN_{\rm ext\_rela}(\mathbf{H}_j^i,\mathbf{H}),{\bm{x}}_j^i})_{j\in \Theta}\to\bm{x},
\label{eq:al}
\end{equation}
where $NN_{\rm ext\_rela}(\cdot)$ refers to the partial network used to extract the relative information among $(\mathbf{H}_j^i,\mathbf{H})$, and $NN_{\rm use\_rela}(\cdot)$ represents another part of the network that couples the extracted relative information with $\bm{x}_j^i$. This design transforms the problem of cross-attribute characterizing $\mathbf{H}_j^i\to \bm{x}_j^i$ into a more tractable intra-attribute interaction problem, namely the characterizing relationships between $\mathbf{H}$ and $\mathbf{H}_j^i$ as well as $\bm{x}$ and $\bm{x}_j^i$, and utilizes dimensionless relative information to achieve cross-attribute transfer, thereby incorporating prior constraints to facilitate learning.

The attention mechanism \cite{attention} provides an effective approach for extracting relative weights among data and guiding the synthesis of new data. Therefore, we design a learning network named Mateformer, whose structure is illustrated in Fig. \ref{fig:mateformer}D. Structurally, Mateformer comprises two Transformer \cite{transformer} modules sharing the same width, referred to as Transformer${}_{\mathbf{H}\bm{x}}$ (the primary Transformer) and Transformer${}_{\mathbf{H}}$ (the auxiliary Transformer), along with several linear layers for dimensional transformation. The inputs to Mateformer are sequences $Seq_{\bm{x}}$ and $Seq_{\mathbf{H}}$, which maintain the element-wise correspondence in order. Here, $Seq_{\bm{x}}$ is composed of $\bm{x}_\mathrm{init}$ and $\bm{x}_j^i~(j\in\Theta)$, while $Seq_{\mathbf{H}}$ consists of $\mathbf{H}$ and $\mathbf{H}_j^i~(j\in\Theta)$. Since the true $\bm{x}$ is to be solved, $\bm{x}_\mathrm{init}$ can be either predefined or learnable parameters. Besides, learnable embeddings are added in Transformer${}_{\mathbf{H}\bm{x}}$ to indicate whether the input is a predefined ${\bm{x}}_\mathrm{init}$ or a known $\bm{x}_j^i$. In this work, we use the mean value of $\bm{x}_j^i$ as ${\bm{x}}_\mathrm{init}$. Moreover, in training, in order to facilitate parallelization, $\mathbf{H}$ can be not just a single sample but a sequence of multiple to-be-inferred samples in the studied scenario. The stacked layers of Mateformer continuously perform interactions and transformations within $Seq_{\bm{x}}$ and $Seq_{\mathbf{H}}$, with the $k$-th layer taking $Seq^k_{\bm{x}}$ and $Seq^k_{\mathbf{H}}$ as inputs and generating $Seq^{k+1}_{\bm{x}}$ and $Seq^{k+1}_{\mathbf{H}}$ as outputs. The final output of Mateformer is the output of $\bm{x}_\mathrm{init}$ from the Transformer${}_{\mathbf{H}\bm{x}}$, namely the inferred location $\widehat{\bm{x}}$.

Each layer of Transformer${}_{\mathbf{H}\bm{x}}$ employs a multi-head attention to implement the relativity-based analogy described in Eq. \ref{eq:al}. The detailed computation flow is shown in Fig. \ref{fig:mateformer}C. First, the relative relationship between $\mathbf{H}$ and $\mathbf{H}_j^i$ is extracted by projecting $Seq^k_{\mathbf{H}}$ into `Query' and `Key' and computing the attention weights. Subsequently, this relativity is used to guide the synthesis of $\bm{x}$ by applying the attention weights to the `Value' projected from $Seq^k_{\bm{x}}$. The masking is applied to the $\mathbf{H}$ and $\bm{x}_\mathrm{init}$ parts of the two sequences, respectively, ensuring that the evaluation of relativity and the synthesis process are only based on the accurate known samples ${\mathbf{H}}_j^i$ and ${\bm{x}}_j^i$. Although the predefined $\bm{x}_\mathrm{init}$ is not directly involved in the attention computation, its information can still propagate forward due to the presence of residual connections shown in Fig. \ref{fig:mateformer}D, enabling the accumulation of synthesized labels in different feature spaces across the stacked layers.

It is important to note that the attention mechanism in Fig. \ref{fig:mateformer}C differs fundamentally from the commonly used cross-attention. Cross-attention aims to fuse information across different data sources or modalities, where one sequence (serving as `Query') selectively attends to relevant information from another sequence (serving as `Key' and `Value'). In contrast, the attention mechanism in the primary Transformer is uniquely designed to extract relative information within a single sequence (serving as `Query' and `Key') and utilizes these correlations to guide the interactions within another sequence (serving as `Value'), which aligns closely with Eq. \ref{eq:al}.

Since each layer of Transformer$_{\mathbf{H}\bm{x}}$ only outputs $Seq^{k+1}_{\bm{x}}$ and does not produce $Seq^{k+1}_{\mathbf{H}}$, we introduce a mate module, Transformer$_{\mathbf{H}}$, to continuously transform the $Seq_{\mathbf{H}}$, thereby generating $Seq^{k+1}_{\mathbf{H}}$ at each layer for the next layer of Transformer${}_{\mathbf{H}\bm{x}}$ to evaluate relativity in another feature space. Consequently, the total number of layers of Transformer${}_{\mathbf{H}}$ will be one less than that of Transformer${}_{\mathbf{H}\bm{x}}$. This design enables stackability of the learning network and maintains the model's deep nonlinear representational capacity. Each layer of Transformer${}_{\mathbf{H}}$ employs multi-head self-attention to enable feature interactions within $Seq^k_{\mathbf{H}}$. The same masking as in Transformer${}_{\mathbf{H}\bm{x}}$ is applied to ensure that attention computation is based on the known data ${\mathbf{H}}_j^i$. 

The overall computation process of Mateformer can be formulated as follows:
\begin{align}
Seq^{k+1}_{\bm{x}}&=\mathrm{FFN}^{k+1}_{\mathbf{H} \bm{x}}(\mathrm{MHead}^{k+1}_{\mathbf{H}\bm{x}}( Seq^k_{\mathbf{H}},Seq^k_{\mathbf{H}},Seq^k_{\bm{x}})),&
k&=1,\dots,N, \label{eq:transformer_Hx}
\\
Seq^{k+1}_{\mathbf{H}}&=\mathrm{FFN}^{k+1}_{\mathbf{H}}(\mathrm{MHead}^{k+1}_{\mathbf{H}}( Seq^k_{\mathbf{H}},Seq^k_{\mathbf{H}},Seq^k_{\mathbf{H}})),&
k&=1,\dots,N-1, \label{eq:transformer_H}
\end{align}
where $\mathrm{FFN}$ is the feedforward network, $\mathrm{MHead}$ is the multi-head attention module, and to highlight the key ideas, the used residual computation and layer normalization are omitted here.

Without introducing the concept of relativity, Eq. \ref{eq:al} would degenerate into Eq. \ref{eq:new_paradigm}, which is similar to in-context learning (ICL) \cite{in_context_survey}, i.e., $NN(\mathbf{H}_{j_1}^i,{\bm{x}}_{j_1}^i,\dots,{\mathbf{H}}_{j_n}^i,{\bm{x}}_{j_n}^i,\mathbf{H} )\to\bm{x}$. Typically, ICL employs a single Transformer model and annotates the correspondence between $\mathbf{H}_j^i$ and $\bm{x}_j^i$ using language descriptions or positional encoding. Such correspondence has to be captured and modeled implicitly during training. In contrast, AL implements explicit structural constraints on the correspondence between $\mathbf{H}_j^i$ and $\bm{x}_j^i$ through Mateformer, a unique bipartite Transformer architecture. Supplementary Text provides detailed comparisons between ICL and AL, illustrating this point.

\subsection*{Experimental validation and property analysis}
We evaluate the accuracy of AL-based localization (ALLoc) and compare it with several representative LNNs, including MFCNet \cite{fdma_positioning}, AD\_CNN \cite{ad_cnn2_long}, and CNN \cite{cnn4}, all of which are the data-to-label methods. Meanwhile, we apply transfer learning \cite{transfer_learning} and multi-task learning \cite{multi_task} on AD\_CNN to evaluate the generalization of ALLoc. Furthermore, given the relevance between ICL and AL, we employ the typical paradigm of ICL to wireless localization (ICLLoc) and compare it with ALLoc. The details about the data-to-label methods and ICLLoc can be found in Supplementary Text.

\subsubsection*{Learning and generalization capability}
\begin{figure}[!htbp]
	\centering
	\includegraphics[width=\textwidth]{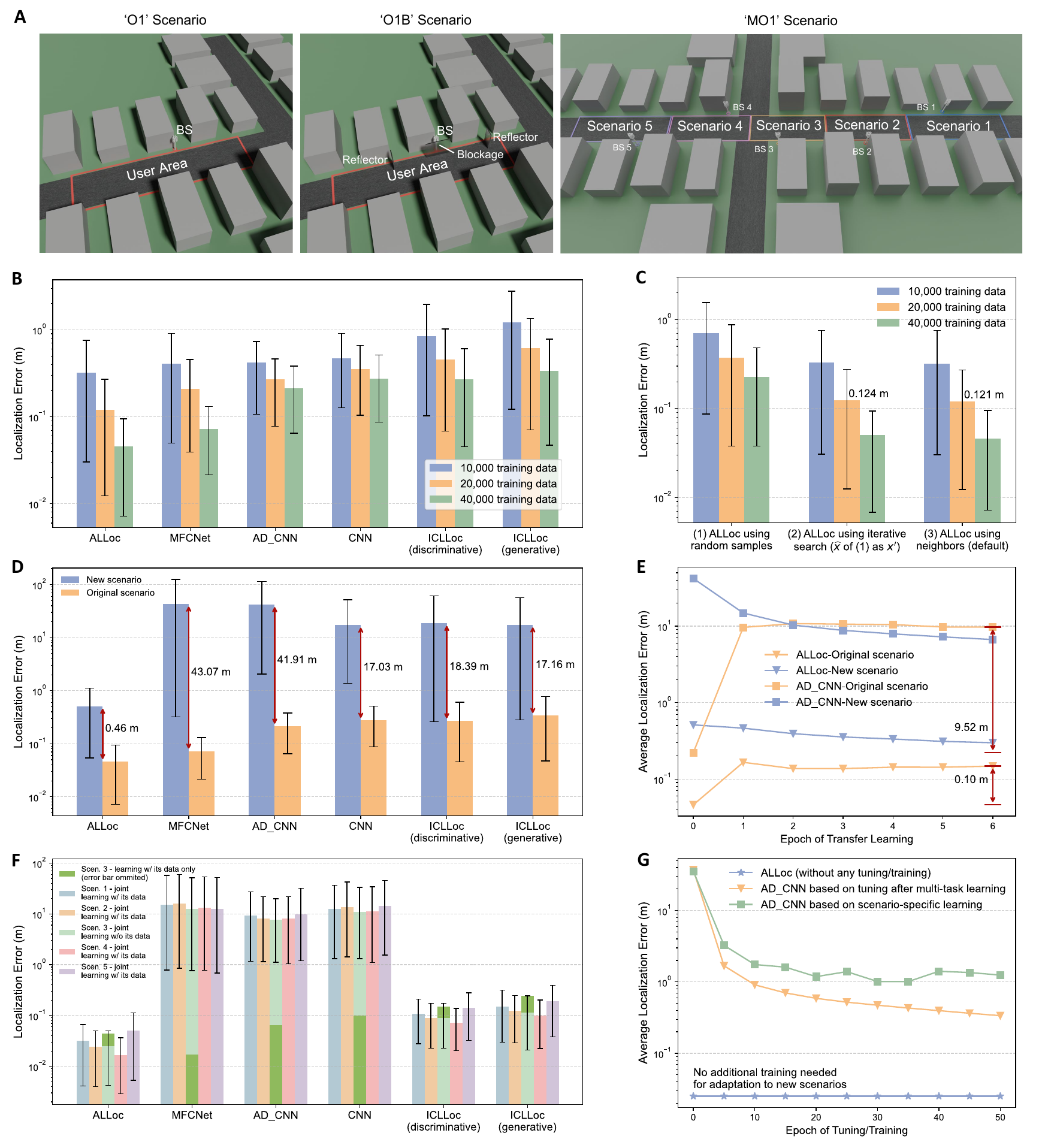}
	\caption{\fontsize{10pt}{10pt}\selectfont \textbf{Experimental results on DeepMIMO dataset, covering single-scenario, cross-scenario, and multi-scenario cases.}
		(\textbf{A}) Aerial view of the experimental scenarios, namely `O1', `O1B', and `MO1'. (\textbf{B}) Localization accuracy in `O1' scenario after training on this scenario. (\textbf{C}) ALLoc's localization accuracy when replacing neighborhood sampling with random sampling or iterative search. (\textbf{D}) Direct reusing the model trained in `O1' scenario to `O1B' scenario. (\textbf{E}) Transfer learning with the model parameters trained in `O1' scenario as initialization and the scenario data of `O1B' scenario as training set. (\textbf{F}) Localization accuracy with multi-scenario joint learning as pre-training, where scenarios 1,2,4, and 5 are involved in joint learning and scenario 3 is excluded. (\textbf{G}) Immediate generalizability in new scenario of pre-trained ALLoc vs. AD\_CNN based on multi-task learning or scenario-specific learning.}
	\label{fig:results_deepmimo}
\end{figure}

To comprehensively analyze the properties of the proposed ALLoc approach, we conduct experiments on the widely-used DeepMIMO dataset \cite{deepmimo}, which is based on ray-tracing simulation \cite{Insite}. Our experiments cover various application modes of single-scenario, cross-scenario, and multi-scenario. The used localization scenarios include the `O1' scenario with consistent line-of-sight (LoS), the `O1B' scenario with non-line-of-sight (NLoS) users due to obstructions, and the multi-scenario `MO1' scenario composed of five adjacent cellulars. Schematic diagrams of the scenarios are shown in Fig. \ref{fig:results_deepmimo}A.

We first present the generalization performance of LNNs in a single scenario after training. Fig. \ref{fig:results_deepmimo}B shows the localization accuracy of ALLoc and baselines in `O1' scenario. Under three different amount of training data, ALLoc achieves accuracy comparable to baselines, with lower average error in most cases. This indicates that the AL architecture based on relativity is not only on par with existing methods but also offers potential improvements in accuracy. Meanwhile, the performance of ICLLoc in single-scenario is modest, even slightly worse than most data-to-label methods. This reflects its limitations in accurate representation and hinders its broader applicability. In Supplementary Text, we conduct more detailed analysis about the robustness of ALLoc against the number of embedded samples, initial location errors, and input channel noise.

Fig. \ref{fig:results_deepmimo}C compares the performance of ALLoc using three different sampling methods: (1) random sampling; (2) iterative search, which uses the inferred result in random sampling serves as the coarse location for neighborhood sampling; and (3) the default neighborhood sampling, which uses a coarse location provided by GNSS. Here, random sampling involves training a new Mateformer adapted for non-neighborhood scenario embedding. The results show that neighborhood sampling achieves higher localization accuracy compared to random sampling, indicating that neighborhood sampling improves the efficiency of scenario information embedding. However, random sampling does not lead to unacceptable performance degradation. Besides, using its result as the coarse value for iterative neighborhood search can achieve performance comparable to that of neighborhood sampling assisted by GNSS. These results demonstrate the effectiveness of neighborhood sampling and also indicate that coarse value assistance from other low-cost localization systems, such as GNSS, can facilitate ALLoc but is not indispensable.

We then demonstrate the cross-scenario generalization capability of LNNs, where the model, after learning within one scenario, is tested in another. Fig. \ref{fig:results_deepmimo}D and fig. \ref{fig:results_deepmimo_o1b_to_o1}A shows the performance of the model trained in `O1' scenario directly deployed to `O1B' scenario and the model trained in `O1B' scenario directly deployed to `O1' scenario. It can be observed that, no matter `O1' to `O1B' or `O1B' to `O1', all of the baseline methods exhibit significant performance degradation, with their localization accuracy completely failing to meet the requirements of wireless localization. In contrast, the proposed ALLoc approach experiences only a slight performance decline. This property demonstrates that ALLoc exhibits excellent robustness to scenario dynamic changes, which is critical for secure and reliable wireless localization. 

Fig. \ref{fig:results_deepmimo}E and fig. \ref{fig:results_deepmimo_o1b_to_o1}B illustrates the immediate generalization performance of ALLoc and AD\_CNN during transfer learning on new scenario data, evaluated in both the new scenario (target) and the original scenario (source). Even after several epochs of transfer learning, the performance of AD\_CNN in the new scenario, while improved, still lags significantly behind the initial performance of ALLoc. Moreover, the performance of ALLoc in the new scenario continues to improve during transfer learning as well. Meanwhile, transfer learning causes catastrophic forgetting for data-to-label methods like AD\_CNN, where adapting to the new scenario leads to a significant performance degradation on the original scenario. This reflects the inability of such methods to reconcile localization intelligence across two different scenarios. In contrast, ALLoc does not suffer from obvious catastrophic forgetting due to transfer learning, and it maintains excellent performance in the original scenario. Under the learning framework of ALLoc, the localization intelligence of two different scenarios can be compatible. These results lead to a hypothesis that the slight decline of ALLoc does not indicate a lack of reusable intelligence across scenarios, but is primarily attributed to the data distribution shift between the new and original scenarios. Furthermore, if ALLoc can be pre-trained on multiple scenarios with diverse data distributions, it is entirely possible for the pre-trained model to operate directly in a new scenario without performance decline. The related experimental results are presented and analyzed in the following.

We demonstrate the multi-scenario generalization capability of LNNs, specifically whether the LNN can efficiently jointly learn in multiple scenario datasets and generalize to new scenarios, akin to the pre-training and deployment process of foundation models. Fig. \ref{fig:results_deepmimo}F shows the performance of LNNs after joint training on scenarios 1, 2, 4, and 5 in `MO1', evaluated on scenarios 1, 2, 4, 5, and the new scenario 3. All of the data-to-label methods fail to achieve effective multi-scenario joint learning. In both the already learned scenarios and the new scenario, their localization errors far exceed those of scenario-specific learning. In contrast, ALLoc achieves highly effective multi-scenario joint learning, not only maintaining high accuracy in the learned scenarios but also delivering superior performance in the new scenario, even surpassing the results of scenario-specific learning. This improvement is attributed to ALLoc's unique architectural design, which eliminates the barriers between multi-scenario localization intelligence and thus facilitates more comprehensive learning by leveraging rich data from diverse scenarios. That is to say, the outcomes of learning the mechanisms of relative analogy are shareable across various scenarios. Inference in a new scenario can be directly founded upon the analogical methods learned in other scenarios. Moreover, ICLLoc is effective as well in generalizing across multiple scenarios, which highlights the universal significance of sample embeddings in helping models understand scenario specificity. However, ALLoc still exhibits a significant performance advantage over ICLLoc, underscoring the   superiority of the AL architecture. 

Fig. \ref{fig:results_deepmimo}G shows the performance of AD\_CNN using multi-task learning and its comparison with ALLoc. Multi-task learning partially leverages the similarities between scenarios to improve the generalizability of the backbone structure, thus achieving faster and more stable convergence compared to scenario-specific learning. However, even after 50 epochs of tuning in the target scenario, its performance still significantly lags behind ALLoc, which requires no additional training. This result further highlights the advantages of the ALLoc in terms of application convenience and localization accuracy.

\subsubsection*{Localization under changing weather and traffic conditions} 
\begin{figure}[!htbp]
	\centering
	\includegraphics[width=\textwidth]{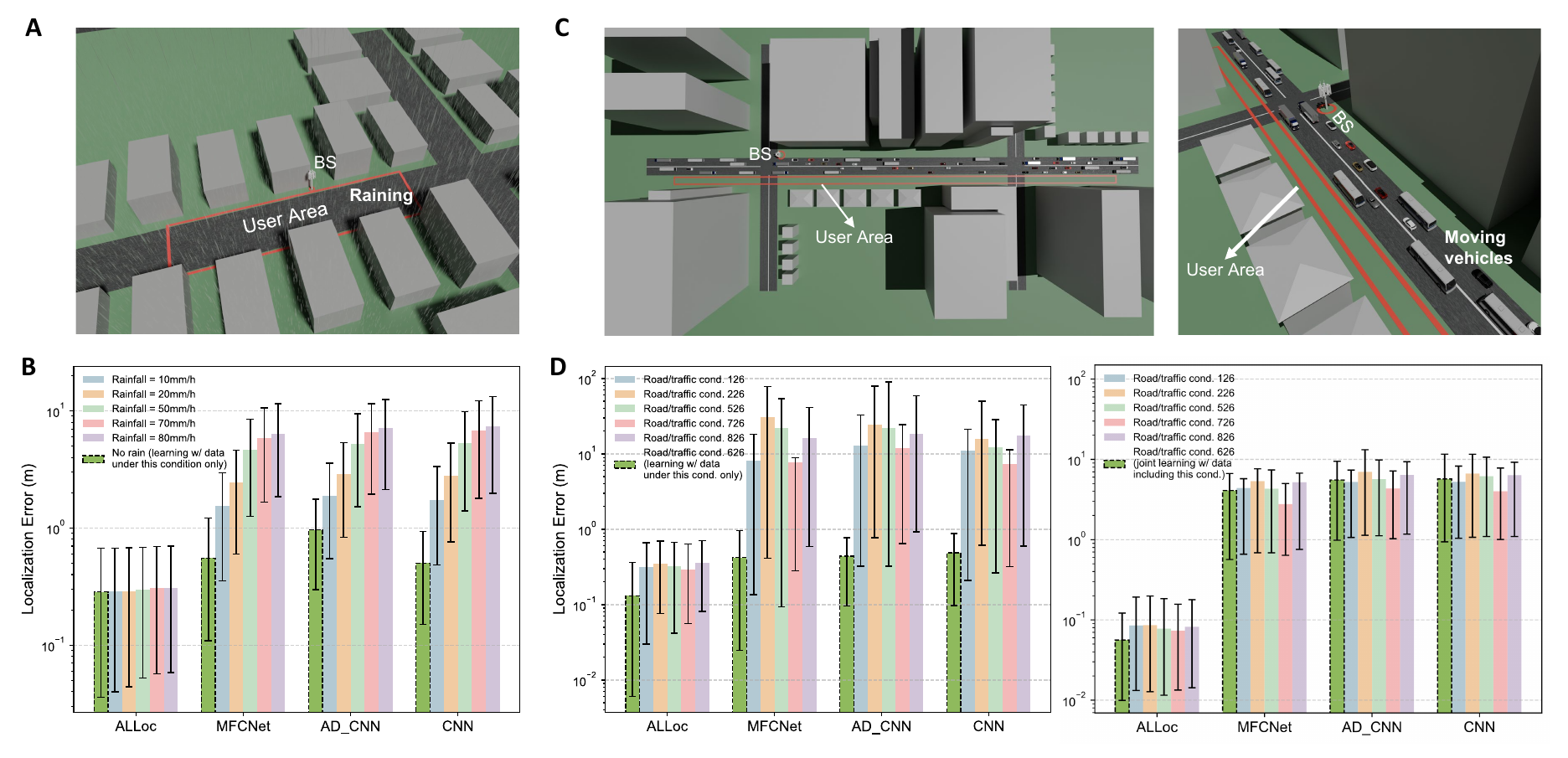}
	\caption{\textbf{Experimental results on DeepMIMO dataset, under different rainfall conditions and dynamic traffic/road conditions.}
		(\textbf{A}) Aerial view of the experimental scenario under rainfall conditions. (\textbf{B}) Reusing the model trained on dry-weather condition to various rainfall scenarios to evaluate the reliability under diverse weather conditions. (\textbf{C}) Overall plan view of the experimental scenario under dynamic road conditions. (\textbf{D}) Reusing the model trained on a single road condition (left) and jointly trained on 45 road conditions (right) to new road conditions. }
	\label{fig:results_deepmimo_2}
\end{figure}

We demonstrate how enhancing generalization can improve localization reliability under complex conditions. We first evaluate the models' robustness to variations in rainfall by training it under dry-weather conditions and testing it under different rainfall intensities. This evaluation was conducted in the `O1' scenario, but certain dataset parameters were modified to highlight the impact of rain attenuation. The modified scenario, as illustrated in Fig. \ref{fig:results_deepmimo_2}A, is referred to as `RO1'. The experimental results are shown in Fig. \ref{fig:results_deepmimo_2}B. It can be observed that MFCNet, AD\_CNN, and CNN these data-to-label methods suffer from performance degradation due to changes in weather, with the degradation becoming more pronounced as rainfall increases. In contrast, ALLoc exhibits minimal performance loss, which can be attributed to its ability to implicitly aware weather changes through embedded pairs from the new scenario and adapt its inference accordingly. This adaptive capability will be helpful in maintaining reliable localization under adverse weather conditions. In a sense, the results in Fig. \ref{fig:results_deepmimo_2}B can be regarded as a concrete manifestation of the cross-scenario generalization capability, now under different rainfall. Furthermore, since rain attenuation affects all multipath components concurrently, albeit to varying extents, the overall structure of the channel data remains relatively stable. Consequently, the performance degradation of ALLoc under varying rainfall conditions is milder than that observed under drastic changes in scattering environments.

In addition, we evaluate the impact of dynamic traffic and road conditions on localization performance. The experiments are conducted using the `O2' scenario from DeepMIMO V2, which is a dynamic road influenced by moving vehicles, as illustrated in Fig. \ref{fig:results_deepmimo_2}C. Fig. \ref{fig:results_deepmimo_2}D shows the localization performance on other road conditions of models trained on a specific road condition and jointly trained on 45 road conditions. When trained on a specific road condition, the accuracy of data-to-label methods deteriorates catastrophically under new road conditions, with errors increasing from the sub-meter level to several tens of meters, rendering them completely unusable. Besides, The high variability among road conditions also hinders the effective joint training of data-to-label methods on multiple road conditions. In contrast, ALLoc trained on a specific road condition consistently maintains sub-meter localization capability in new road conditions, which markedly enhances the reliability of wireless localization under complex traffic conditions. Furthermore, through joint training across diverse road conditions, ALLoc not only learns effectively but also enhances its generalization capability, achieving performance on new conditions that surpasses models trained and tested on the specific condition. Such an excellent characteristic makes it a robust tool for high-reliability wireless localization amidst challenging road conditions. In a sense, the results in Fig. \ref{fig:results_deepmimo_2}D is a manifestation of the cross-scenario and multi-scenario generalization capability, within the representative application of dynamic traffic. Firstly, due to its robust transferability, an ALLoc model trained in a specific scenario retains its basic operational capacity in other scenarios with only minor performance degradation. Secondly, by leveraging a learning architecture that accommodates inter-scenario variability, ALLoc permits the joint training of a single model on richer, multi-scenario datasets, which yields superior inference accuracy and broader generalization.

\subsubsection*{Validation on real-world data}
\begin{figure}[!htbp]
	\centering
	\includegraphics[width=\textwidth]{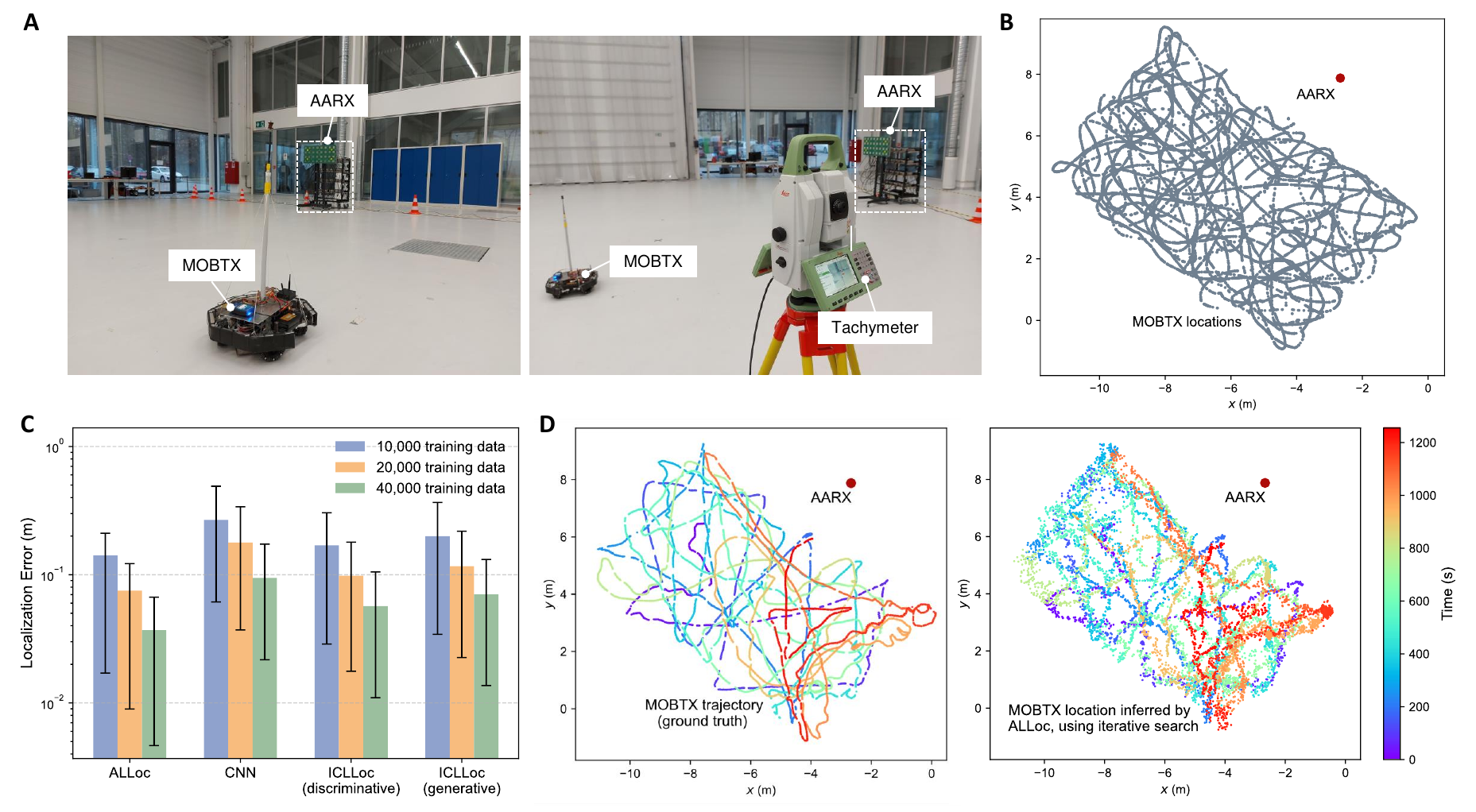}
	\caption{\textbf{Experimental results on DICHASUS-ca0x dataset.}
		(\textbf{A}) Measurement setup in the indoor mobile communication scenario. Photos adopted from \cite{dichasus-ca0x}. (\textbf{B}) MOBTX trajectory ca01, part of which was used to construct the training dataset. (\textbf{C}) Localization accuracy on the remaining part of ca01. (\textbf{D}) Using ALLoc to infer MOBTX locations on a new trajectory ca02.}
	\label{fig:results_dichasus}
\end{figure}

To validate the effectiveness of ALLoc in practical wireless systems, we conduct experiments on DICHASUS dataset \cite{dichasus}, an open-source large-scale MIMO CSI dataset developed by the University of Stuttgart. DICHASUS contains real-world CSI measurement data annotated with location coordinates and timestamps. Here, we use the DICHASUS-ca0x subset \cite{dichasus-ca0x}, which is collected in the indoor mobile communication scenario shown in Fig. \ref{fig:results_dichasus}A. The antenna array receiver (AARX) simulates the BS, while the mobile transmitter (MOBTX) with one transmitting antenna resembles the user equipment. A tachymeter is used to precisely track the MOBTX location. The ca0x subset consists of channel data collected along two trajectories of the MOBTX, namely ca01 and ca02. We select a portion of locations from the ca01 trajectory to construct the training dataset. The remaining portion of ca01 and the complete new trajectory ca02 are both used for evaluation. The discrepancy between ca01 and ca02 reflects real-world data collection situations: since user trajectories used in training are unlikely to cover the entire activity area perfectly, it is typically inevitable to encounter new user trajectories during deployment, where the user location has no closely neighboring reference points in the training dataset. Therefore, we incorporate ca02 into evaluation to further investigate the feasibility of ALLoc in real-world systems.

We first evaluate the localization accuracy of LNNs on the ca01 trajectory, where the training and testing dataset are highly consistent. Fig. \ref{fig:results_dichasus}B visualizes the sampling points along the ca01 trajectory. Given that the coarse location $\bm{x}'$ provided by GNSS is typically unavailable in such indoor scenarios, we apply iterative search to construct the neighborhood, i.e., using the inferred result by random sampling as the coarse location for new neighborhood sampling. Fig. \ref{fig:results_dichasus}C shows the localization accuracy of ALLoc compared to several baseline models. Across three different training data sizes, ALLoc achieves accuracy comparable to or superior to the baseline models not only in the aforementioned simulation experiments but also on real-world data, validating the practicality and superiority of the proposed ALLoc approach.

Next, we apply ALLoc to perform localization on the new trajectory, ca02, with the inference process still incorporating embedded samples from the previous training dataset based on ca01. Fig. \ref{fig:results_dichasus}D shows the true trajectory of ca02 alongside the trajectory formed by the locations inferred by ALLoc. 
The location points are assigned different colors based on the timestamps of the channel measurements corresponding to these locations. This prevents biased localization results from being confused with other true nearby locations, thereby more clearly demonstrating the trajectory tracking accuracy of ALLoc. 
Clearly, most of the inferred locations are distributed near the true trajectory, forming a continuous movement curve with the correct color gradient. In summary, these results validate the effectiveness of ALLoc in real indoor scenarios where GNSS assistance is unavailable.

\subsubsection*{Demonstration of city-scale applications}
Finally, we conduct experiments to validate ALLoc’s cross-scenario generalization capability under more realistic urban settings. To this end, we collect the 3D geographic data of three global cities, namely San Francisco, Shanghai, and Singapore, from the OpenStreetMap database \cite{openstreetmap}, and manually construct 3 scenarios for each city, yielding a total of 9 scenarios. Then we employ Sionna ray-tracing platform \cite{sionna,sionnart} to simulate radio wave propagation and generate channel data.

\begin{figure}[!htbp]
	\centering
	\includegraphics[width=\textwidth]{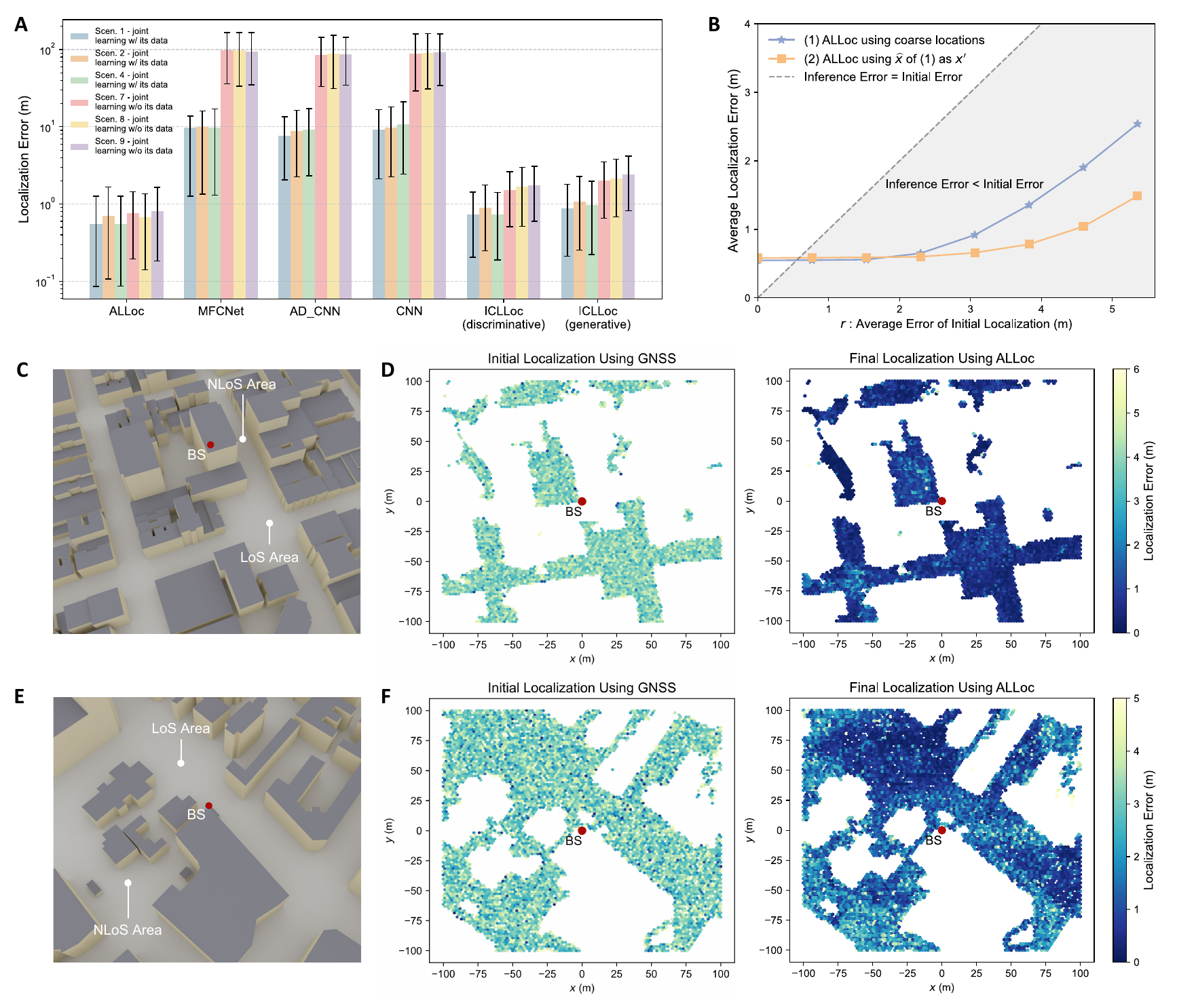}
	\caption{\textbf{Experimental results on realistic urban scenarios.}
		(\textbf{A}) Localization accuracy with multi-scenario joint learning as pre-training, where scenarios 1 to 6 (San Francisco and Shanghai) are involved in joint learning and scenarios 7 to 9 (Singapore) are excluded. (\textbf{B}) ALLoc’s localization accuracy versus accuracy of initial location error. (\textbf{C} and \textbf{E}) Aerial views of scenario 1 (C) and scenario 9 (E). (\textbf{D} and \textbf{F}) Spatial distribution of initial location errors and ALLoc's inference errors in scenario 1 (D) and scenario 9 (F).}
	\label{fig:results_sionna}
\end{figure}

We train the LNNs jointly using datasets collected from scenarios 1 to 6 (scenarios in San Francisco and Shanghai) and evaluate on all scenarios, including the remaining unseen scenarios 7 to 9 (scenarios in Singapore). This demonstrates a typical application paradigm of wireless localization foundation models: the model is pre-trained using limited data collected across several cellular scenarios, and then it can be directly deployed in new cities worldwide to perform wireless localization with the assistance of local databases. For such city-scale outdoor scenarios, we leverage the coarse location provided by GNSS as available auxiliary information to improve localization efficiency. Fig. \ref{fig:results_sionna}A shows the localization accuracy of LNNs. The data-to-label methods, MFCNet, AD\_CNN, and CNN, all fail to achieve effective multi-scenario joint learning. Moreover, due to the significant differences among the scenarios, the accuracy of these methods further declines in the new scenarios, with localization errors increasing from an already unacceptable ten-meter level to nearly 100 meters. Although ICLLoc performs well in the learned scenarios, its performance degrades when evaluated in new scenarios, with errors reaching several meters. This stems from the fact that ICLLoc lacks a structural prior design, resulting in inefficient utilization of scenario embeddings. In contrast, ALLoc maintains an average localization accuracy of sub-meter level in both the learned scenarios and new scenarios, highlighting the superiority of relativity-based architecture design.

Fig. \ref{fig:results_sionna}B illustrates the relationship between ALLoc’s localization accuracy and the initial coarse location accuracy, using scenario 1 as an illustrative example. When the initial error is less than 2\,$\text{m}$, the final localization performance remains stable without significant degradation, even as the initial error increases. When the initial error exceeds 2\,$\text{m}$, the localization error increases as the initial error grows. Nevertheless, the localization error remains consistently lower than the initial coarse location error, validating ALLoc’s ability for location refinement. This observation leads to a new iterative search strategy: when the coarse location $\bm{x}'$ has a large deviation, the result inferred through neighborhood sampling still exhibits mediocre performance; in such cases, the inferred location $\widehat{\bm{x}}$ can serve as a new initial location for a second round of neighborhood sampling, thereby iteratively improving the localization accuracy. We employ this multi-step neighborhood sampling approach in scenarios 1 and 9, which serve as examples of the already learned scenarios and the new scenarios, respectively. The aerial views of these two scenarios are presented in Figs. \ref{fig:results_sionna}C and \ref{fig:results_sionna}E, where the dense buildings create NLoS areas, resulting in extremely complex channel characteristics that pose challenges for intelligent wireless localization. Figs. \ref{fig:results_sionna}D and \ref{fig:results_sionna}F visualize the spatial distribution of ALLoc's localization errors. In most areas of both scenarios, ALLoc reduces GNSS errors from several meters to less than one meter, demonstrating more precise localization ability.

\section*{Discussion}
Cross-scenario generalization remains a persistent challenge in deep learning, stemming from the tendency of data-driven learning processes to severely overfit the specific scenarios from which the data originates. Existing approaches, such as meta-learning and transfer learning, essentially manage, rather than eliminate, the scenario-specificity of the learned models. This work first points out that this limitation is rooted in an over-reliance on absolute numerical values during feature representation. Consequently, we introduce relativity—a foundational concept in physics—into representation learning, and subsequently propose the AL approach. By introducing reference anchors and performing relativity-based analogical reasoning, the AL architecture generalizes across multiple scenarios, offering a robust solution for learning in system-level wireless localization, as well as for more multi-platform and multi-spatiotemporal systems.

On the other hand, the expansion of intelligence necessitates an additional focusing step during the inference phase, thereby introducing extra complexity. Specifically, as the process shifts from inferring a single sample to simultaneously processing $n$ samples, the complexity, taking Mateformer as an example, increases by a factor of $\mathcal{O}(n^2)$. Consequently, this gives rise to new research directions regarding the trade-off between the scope of intelligence and inference focus. Optimizing this involves developing more streamlined computational architectures and refining context embedding mechanisms to mitigate current quadratic increase in complexity.

Finally, from the perspective of foundation model pre-training and adaptation, the embedding of data-label pairs can be seen as analogous to few-shot prompts for scenario information. However, multi-attribute natural signals such as CSI and location lack the inherent `stickiness' of language -- they are not pre-processed by human intelligence and thus not reasonable to be combined in a context-aware manner like text \cite{wnbaim}. As a result, handling such structured prompts requires more than a simple sequence model; it necessitates an architecture with dedicated `interfaces', such as the Mateformer proposed in this article, to enable efficient reading and utilization of these natural signals. The structural design of Mateformer may thus offer valuable insights for the development of foundation models in scientific and engineering-related fields.

\section*{Materials and Methods}
\subsection*{Dataset construction}
All of our experiments consider the typical multiple-input multiple-output orthogonal frequency division multiplexing (MIMO-OFDM) system, where the BS is equipped with $N_{\rm t}$ antennas and employs OFDM modulation with $N_{\rm c}$ subcarriers for communication with single-antenna user equipments. Mathematically, the wireless channel $\mathbf{H}$ can be characterized by a complex-valued frequency response matrix, i.e., $\mathbf{H}\in\mathbb{C}^{N_{\rm t}\times N_{\rm c}}$. In our experiments, we set the number of antennas as $N_{\rm t}=32$ and the number of subcarriers as $N_{\rm c}=32$.

The experiments in `O1', `O1B', and `MO1' scenarios are based on DeepMIMO dataset (V1), while the experiments in `O2' scenario are based on DeepMIMO dataset (V2). Data samples with zero or excessively small channel amplitudes have been filtered out, as such samples cannot be collected in real-world systems. In all these scenarios, the BS employs a uniform linear antenna array, and channel data is generated at a frequency of 3.5 GHz with a bandwidth of 40 MHz. In this paper, the `O2' scenario involves 50 road conditions, namely cond. 6, 26, 46, $\dots$, 996. The `RO1' scenario builds upon the `O1' scenario by incorporating rainfall conditions. Since rain attenuation is particularly pronounced for high-frequency electromagnetic signals, we regenerated the channel data using a 60 GHz carrier frequency and a 1.6 GHz bandwidth, while keeping all other parameters consistent with the `O1' scenario. Rain attenuation was synthesized based on the measurement results from international telecommunication union \cite{ITU-R-P838-3}. More detailed parameters are provided in table \ref{tab:deepmimo}.

The validation on real-world data is based on DICHASUS-ca0x dataset \cite{dichasus,dichasus-ca0x}. The channel coefficients were measured by a $4\times 8$ uniform planar antenna array across 1,024 subcarriers, with a carrier frequency of 1.272 GHz and a bandwidth of 50 MHz. To reduce the computational and memory requirements for data processing, we downsampled the frequency-domain signals by a factor of 32 to obtain channel data on 32 subcarriers. We then randomly selected 60,000 and 10,000 data samples from the ca01 and ca02 trajectories, respectively. Of these, 40,000 data samples from ca01 are used to form the training dataset, while the remaining samples (including 20,000 from ca01 and 10,000 from ca02) are used for evaluation.

To conduct experiments in realistic urban scenarios, we collect the 3D geographic data of several global cities, namely San Francisco, Shanghai, and Singapore, from the OpenStreetMap database \cite{openstreetmap}. For each city, we manually set three BS locations on top of the roofs of certain buildings, constructing a total of 9 scenarios. The satellite images of these three cities and the 9 constructed scenarios are shown in fig. \ref{fig:map}. These 3D city representations are then imported into Sionna ray-tracing platform \cite{sionna,sionnart} to simulate radio wave propagation and generate channel data. In each scenario, the users are randomly distributed within a 200 m $\times$ 200 m square area centered on the BS. The parameters for data generation, including carrier frequency, bandwidth, BS antenna array form, and training/testing dataset size, are consistent with the `MO1' scenario.

\subsection*{Performance index}
We use the Euclidean distance between the inferred location and the true location as the localization error, defined as follows:
\begin{equation}
\mathrm{Error}=\|\bm{x}-\widehat{\bm{x}}\|_2,
\end{equation}
where $\bm{x}$ is the true value, $\widehat{\bm{x}}$ is the inferred value. In all the experimental result figures in this paper, the colored bars represent the mean localization error over all data in the testing dataset, while the error bars indicate the range of errors from the 10th to the 90th percentile of the sorted error distribution. 

\subsection*{Training and testing of AL}
Data-to-label methods directly utilize CSI-location ($\mathbf{H}$-$\bm{x}$) pairs for training and testing. Differently, AL requires leveraging these pairs to generate neighborhood sequences. Specifically, for each $\mathbf{H}$-$\bm{x}$ pair, AL first searches the training set of the studied scenario to identify the $n$-nearest points $\bm{x}_j$ and their corresponding $\mathbf{H}_j$, forming the neighborhood. This process generates paired sets: the $\{\mathbf{H}\}$ set and the $\{\bm{x}\}$ set, each containing $n+1$ elements (such as $\bm{x}$ and its $n$-nearest neighbors). During each training step, AL performs two sampling operations from these $n+1$ element-pairs: the first randomly selects $p$ data-label pairs for neighborhood embedding, and the second randomly selects $q$ pairs as the to-be-inferred samples. Besides, these two samplings do not require completely different samples, as Mateformer's prior design ensures that the model relies on relativity weight for inference. Thus, even if overlapping samples exist in the sampling process, the training still obtains effective supervision. The training of Mateformer is based on the MSE loss function and optimized using a gradient descent-based optimizer, with Adam \cite{adam} being employed in this work. Detailed information on parameters such as $n$, $p$, and $q$, as well as the training settings, can be found in table \ref{tab:model}.

During the inference/testing phase, the coarse initial location $\bm{x}'$ is generated as follows:
\begin{equation}
    \bm{x}'=\bm{x}+\Delta\bm{x},\quad\Delta\bm{x}=[c,d],
\label{eq:coarse_loc}
\end{equation}
where $c$ and $d$ obey to the uniform distribution over $[-l,l]$, with $l=1\,\text{m}$ by default. This setting is according to the state-of-the-art GPS localization accuracy \cite{gps_accuracy}. The average error of initial location can be expressed as
\begin{equation}
    r=\frac{1}{3}(\sqrt{2}+\ln(1+\sqrt{2}))\cdot l, \label{eq:init_loc_error}
\end{equation}
which is the mathematical expectation of $\|\Delta\bm{x}\|_2$.

In inference, AL also searches the training set of the studied scenario to identify the $n$-nearest points $\bm{x}_j$ around $\bm{x}'$ and their corresponding $\mathbf{H}_j$. It then combines these neighbors with $\mathbf{H}$ to infer $\widehat{\bm{x}}$. Besides, AL can also adopt random sampling without initial locations, by simply replaces the neighborhood search with random sampling during training and testing, while keeping the rest unchanged. The inferred result by random sampling can serve as the initial location for new neighborhood sampling.

In indoor scenarios (such as DICHASUS-ca0x), the situation is slightly different. First, the coarse location provided by GNSS is typically unavailable in indoor scenarios. Hence, during inference, random sampling has to be employed to infer an initial location for new neighborhood sampling. Second, since the indoor user area is typically small, the spatial distribution of training data is extremely high in density. In this case, the $n$-nearest points selected from the training set almost always coincide with the initial location, making it unlikely to provide effective scenario characteristics for localization, especially when the initial location is inaccurate. Therefore, for neighborhood sampling, we randomly select data points within a one-meter radius from the initial location to form the neighborhood.

\newpage




\clearpage 

%
\bibliography{references} 

@article{wBAIM,
  title={Big {AI} models for {6G} wireless networks: Opportunities, challenges, and research directions}, 
  author={Chen, Zirui and Zhang, Zhaoyang and Yang, Zhaohui},
  journal={IEEE Wireless Communications}, 
  volume={31},
  number={5},
  pages={164-172},
  year={2024}
}

@misc{5g-base-stations,
  author={{China Data Portal (2026)}},
  title={{5G Base Stations: China vs World}},
  url={https://chinadata.live/data/5g-base-stations-china-vs-world/},
  note={Accessed: 2026}
}

@misc{gps_accuracy,
  author={{U.S. Department of Transportation}},
  title={{GPS} accuracy},
  url={https://www.gps.gov/systems/gps/performance/accuracy/},
  note={Accessed: 2025}
}

@article{deepmimo,
  title={{DeepMIMO}: A generic deep learning dataset for millimeter wave and massive {MIMO} applications},
  author={Alkhateeb, A.},
  journal={arXiv preprint arXiv:1902.06435},
  year={2019}
}

@Misc{Insite,
  author = {{Remcom}},
  title = {Wireless InSite},
  url = {http://www.remcom.com/wireless-insite},
  note = {Accessed: 2019}, 
}

@inproceedings{transformer,
 title={Attention is all you need},
 author={Vaswani, A. and Shazeer, N. and Parmar, N. and others},
 booktitle={Advances in Neural Information Processing Systems},
 volume={30},
 pages={},
 year={2017}
}

@InProceedings{pre_norm,
  title={On layer normalization in the {Transformer} architecture},
  author={Xiong, R. and Yang, Y. and He, D. and others},
  booktitle={Proceedings of the 37th International Conference on Machine Learning},
  volume={119},
  year={2020},
}

@article{attention,
  title={Neural machine translation by jointly learning to align and translate},
  author={Bahdanau, D. and Cho, K. and Bengio, Y.},
  journal={arXiv preprint arXiv:1409.0473},
  year={2014}
}

@inproceedings{autonomous_driving_2,
  title={Domain adaptive object detection for autonomous driving under foggy weather},
  author={Li, Jinlong and Xu, Runsheng and Ma, Jin and Zou, Qin and Ma, Jiaqi and Yu, Hongkai},
  booktitle={2023 IEEE/CVF Winter Conference on Applications of Computer Vision (WACV)},
  pages={612--622},
  year={2023}
}

@article{robot_manipulation,
  title={Trends and challenges in robot manipulation},
  author={Billard, Aude and Kragic, Danica},
  journal={Science},
  volume={364},
  number={6446},
  pages={eaat8414},
  year={2019}
}

@inproceedings{robot_manipulation_2,
  title={Transferring foundation models for generalizable robotic manipulation},
  author={Yang, Jiange and Tan, Wenhui and Jin, Chuhao and Yao, Keling and Liu, Bei and Fu, Jianlong and Song, Ruihua and Wu, Gangshan and Wang, Limin},
  booktitle={2025 IEEE/CVF Winter Conference on Applications of Computer Vision (WACV)},
  pages={1999--2010},
  year={2025}
}

@article{fdma_positioning,
  title={Deep learning-based multi-user positioning in wireless {FDMA} cellular networks}, 
  author={Chen, Z. and Zhang, Z. and Xiao, Z. and others},
  journal={IEEE Journal on Selected Areas in Communications}, 
  volume={41},
  number={12},
  pages={3848-3862},
  year={2023},
}

@inproceedings{ad_cnn1,
  title={Deep convolutional neural networks for massive {MIMO} fingerprint-based positioning}, 
  author={Vieira, J. and Leitinger, E. and Sarajlic, M. and Li, X. and Tufvesson, F.},
  booktitle={2017 IEEE 28th Annual International Symposium on Personal, Indoor, and Mobile Radio Communications (PIMRC)}, 
  volume={},
  number={},
  pages={1-6},
  year={2017},
}

@inproceedings{ad_cnn2,
  title={Deep convolutional neural networks enabled fingerprint localization for massive {MIMO-OFDM} system}, 
  author={Sun, X. and Wu, C. and Gao, X. and Li, G.Y.},
  booktitle={2019 IEEE Global Communications Conference (GLOBECOM)}, 
  volume={},
  number={},
  pages={1-6},
  year={2019},
}

@article{ad_cnn2_long,
  title={Fingerprint-based localization for massive {MIMO-OFDM} system with deep convolutional neural networks}, 
  author={Sun, X. and Wu, C. and Gao, X. and Li, G.Y.},
  journal={IEEE Transactions on Vehicular Technology},
  volume={68},
  number={11},
  pages={10846-10857}, 
  year={2019},
}

@article{ad_cnn3_long,
  title={Learning to localize: A {3D} {CNN} approach to user positioning in massive {MIMO-OFDM} systems}, 
  author={Wu, C. and Yi, X. and Wang, W. and You, L. and Huang, Q. and Gao, X. and Liu, Q.},
  journal={IEEE Transactions on Wireless Communications},
  volume={20},
  number={7},
  pages={4556-4570}, 
  year={2021},
}

@inproceedings{cnn1,
  title={Novel massive {MIMO} channel sounding data applied to deep learning-based indoor positioning}, 
  author={Arnold, M. and Hoydis, J. and Brink, S.T.},
  booktitle={SCC 2019; 12th International ITG Conference on Systems, Communications and Coding}, 
  volume={},
  number={},
  pages={1-6},
  year={2019},
}

@inproceedings{cnn2,
  title={DelFin: A deep learning based {CSI} fingerprinting indoor localization in {IoT} context}, 
  author={Berruet, B. and Baala, O. and Caminada, A. and Guillet, V.},
  booktitle={2018 International Conference on Indoor Positioning and Indoor Navigation (IPIN)}, 
  volume={},
  number={},
  pages={1-8},
  year={2018},
}

@inproceedings{cnn4,
  title={{CSI}-based positioning in massive {MIMO} systems using convolutional neural networks}, 
  author={Bast, S.D. and Guevara, A.P. and Pollin, S.},
  booktitle={2020 IEEE 91st Vehicular Technology Conference (VTC2020-Spring)}, 
  volume={},
  number={},
  pages={1-5},
  year={2020},
}

@article{adam,
  title={Adam: A method for stochastic optimization},
  author={Kingma, D.P. and Ba, J.},
  journal={arXiv preprint arXiv:1412.6980},
  year={2014}
}

@article{channel_deduction,
  title={Channel deduction: A new learning framework to acquire channel from outdated samples and coarse estimate}, 
  author={Chen, Zirui and Zhang, Zhaoyang and Yang, Zhaohui and Huang, Chongwen and Debbah, Mérouane},
  journal={IEEE Journal on Selected Areas in Communications}, 
  volume={},
  number={},
  pages={1-1},
  year={2025},
}

@article{loc_importance2,
  title={Deep learning-based robust positioning for all-weather autonomous driving},
  author={Almalioglu, Yasin and Turan, Mehmet and Trigoni, Niki and others},
  journal={Nature Machine Intelligence},
  volume={4},
  number={9},
  pages={749--760},
  year={2022},
}

@book{gnss,
  author={Peter J.G. Teunissen and Oliver Montenbruck},
  title={Springer Handbook of Global Navigation Satellite Systems},
  series={Springer Handbooks},
  publisher={Springer Cham},
  year={2017},
}

@INPROCEEDINGS{transloc1,
  author={Li, Peizheng and Cui, Han and Khan, Aftab and Raza, Usman and Piechocki, Robert and Doufexi, Angela and Farnham, Tim},
  booktitle={2021 IEEE Radar Conference (RadarConf21)}, 
  title={Deep transfer learning for {WiFi} localization}, 
  year={2021},
  volume={},
  number={},
  pages={1-5},
}

@ARTICLE{transloc2,
  author={Li, Lin and Guo, Xiansheng and Zhao, Mengxue and Li, Huiyong and Ansari, Nirwan},
  journal={IEEE Transactions on Wireless Communications}, 
  title={{TransLoc}: A heterogeneous knowledge transfer framework for fingerprint-based indoor localization}, 
  year={2021},
  volume={20},
  number={6},
  pages={3628-3642},
}

@ARTICLE{metaloc,
  author={Gao, Jun and Wu, Dongze and Yin, Feng and Kong, Qinglei and Xu, Lexi and Cui, Shuguang},
  journal={IEEE Journal on Selected Areas in Communications}, 
  title={{MetaLoc}: Learning to learn wireless localization}, 
  year={2023},
  volume={41},
  number={12},
  pages={3831-3847},
}

@ARTICLE{multi_task,
  author={Zhang, Yu and Yang, Qiang},
  journal={IEEE Transactions on Knowledge and Data Engineering}, 
  title={A survey on multi-task learning}, 
  year={2022},
  volume={34},
  number={12},
  pages={5586-5609},
}

@ARTICLE{multi_task_csinet,
  author={Li, Xiangyi and Guo, Jiajia and Wen, Chao-Kai and Jin, Shi and Han, Shuangfeng and Wang, Xiaoyun},
  journal={IEEE Transactions on Communications}, 
  title={Multi-task learning-based {CSI} feedback design in multiple scenarios}, 
  year={2023},
  volume={71},
  number={12},
  pages={7039-7055},
}

@ARTICLE{transfer_learning,
  author={Zhuang, Fuzhen and Qi, Zhiyuan and Duan, Keyu and Xi, Dongbo and Zhu, Yongchun and Zhu, Hengshu and Xiong, Hui and He, Qing},
  journal={Proceedings of the IEEE}, 
  title={A comprehensive survey on transfer learning}, 
  year={2021},
  volume={109},
  number={1},
  pages={43-76},
}

@ARTICLE{cooper_loc,
  author={Wymeersch, Henk and Lien, Jaime and Win, Moe Z.},
  journal={Proceedings of the IEEE}, 
  title={Cooperative localization in wireless networks}, 
  year={2009},
  volume={97},
  number={2},
  pages={427-450},
}

@article{wireless_loc_survey,
  title={Positioning using wireless networks: Applications, recent progress and future challenges},
  author={Yang, Yang and Chen, Mingzhe and Blankenship, Yufei and Lee, Jemin and Ghassemlooy, Zabih and Cheng, Julian and Mao, Shiwen},
  journal={arXiv preprint arXiv:2403.11417},
  year={2024}
}

@INPROCEEDINGS{mccnet,
  author={Chen, Zirui and Zhang, Zhaoyang and Xiao, Zhuoran and Zhang, Chuanzhi and Yang, Zhaohui},
  booktitle={2023 IEEE 34th Annual International Symposium on Personal, Indoor and Mobile Radio Communications (PIMRC)}, 
  title={{CSI} of each subcarrier is a fingerprint: Multi-carrier cumulative learning based positioning in massive {MIMO} systems}, 
  year={2023},
  volume={},
  number={},
  pages={1-7},
}

@inproceedings{in_context_survey,
    title = "A survey on in-context learning",
    author = "Dong, Qingxiu  and
      Li, Lei  and
      Dai, Damai  and
      Zheng, Ce  and
      Ma, Jingyuan  and
      Li, Rui  and
      Xia, Heming  and
      Xu, Jingjing  and
      Wu, Zhiyong  and
      Chang, Baobao  and
      Sun, Xu  and
      Li, Lei  and
      Sui, Zhifang",
    booktitle = "Proceedings of the 2024 Conference on Empirical Methods in Natural Language Processing",
    year = "2024",
    pages = "1107--1128",
}

@ARTICLE{meta_learning,
  author={Hospedales, Timothy and Antoniou, Antreas and Micaelli, Paul and Storkey, Amos},
  journal={IEEE Transactions on Pattern Analysis and Machine Intelligence}, 
  title={Meta-learning in neural networks: A survey}, 
  year={2022},
  volume={44},
  number={9},
  pages={5149-5169},
}

@ARTICLE{domain_generalization,
  author={Wang, Jindong and Lan, Cuiling and Liu, Chang and Ouyang, Yidong and Qin, Tao and Lu, Wang and Chen, Yiqiang and Zeng, Wenjun and Yu, Philip S.},
  journal={IEEE Transactions on Knowledge and Data Engineering}, 
  title={Generalizing to unseen domains: A survey on domain generalization}, 
  year={2023},
  volume={35},
  number={8},
  pages={8052-8072},
}

@ARTICLE{domain_adaptation,
  author={Guan, Hao and Liu, Mingxia},
  journal={IEEE Transactions on Biomedical Engineering}, 
  title={Domain adaptation for medical image analysis: A survey}, 
  year={2022},
  volume={69},
  number={3},
  pages={1173-1185},
}

@article{in_context_estimator,
  title={{Transformers} are provably optimal in-context estimators for wireless communications},
  author={Teja Kunde, Vishnu and Rajagopalan, Vicram and Shekhara Kaushik Valmeekam, Chandra and Narayanan, Krishna and Shakkottai, Srinivas and Kalathil, Dileep and Chamberland, Jean-Francois},
  journal={arXiv e-prints},
  pages={arXiv--2311},
  year={2023}
}

@INPROCEEDINGS{in_context_equalization,
  author={Zecchin, Matteo and Yu, Kai and Simeone, Osvaldo},
  booktitle={2024 IEEE International Conference on Communications Workshops (ICC Workshops)}, 
  title={In-context learning for {MIMO} equalization using {Transformer}-based sequence models}, 
  year={2024},
  volume={},
  number={},
  pages={1573-1578},
}

@article{wnbaim,
  title={Towards wireless native big {AI} model: The mission and approach differ from large language model},
  author={Chen, Zirui and Zhang, Zhaoyang and Liu, Chenyu and Xing, Ziqing},
  journal={Science China Information Sciences},
  volume={68},
  number={7},
  pages={170303},
  year={2025}
}

@book{dl_book,
  title={Deep learning},
  author={Goodfellow, Ian and Bengio, Yoshua and Courville, Aaron},
  year={2016},
  publisher={MIT Press},
  series={Adaptive Computation and Machine Learning series}
}

@techreport{ITU-R-P838-3,
  author={{International Telecommunication Union - Radiocommunication Sector}},
  title={{Recommendation {ITU-R} P.838-3: Specific attenuation model for rain for use in prediction methods}},
  institution={International Telecommunication Union},
  year={2005},
  url={https://www.itu.int/rec/r-rec-p.838-3-200503-i}
}

@inproceedings{dichasus,
  author={Florian Euchner and Marc Gauger and Sebastian D\"orner and Stephan ten Brink},
  title={A distributed massive {MIMO} channel sounder for “big {CSI} data”-driven machine learning}, 
  booktitle={25th International ITG Workshop on Smart Antennas (WSA)}, 
  year={2021},
  volume={},
  number={},
  pages={1-6},
}

@data{dichasus-ca0x,
	author={Euchner, Florian and Gauger, Marc},
	publisher={DaRUS},
	title={{CSI Dataset dichasus-ca0x: Industrial Environment LoS Day 1}},
	url={https://dichasus.inue.uni-stuttgart.de/datasets/data/dichasus-ca0x/},
	year={2022}
}

@article{sionna,
  title={Sionna: An open-source library for next-generation physical layer research},
  author={Hoydis, Jakob and Cammerer, Sebastian and Aoudia, Fay{\c{c}}al Ait and Vem, Avinash and Binder, Nikolaus and Marcus, Guillermo and Keller, Alexander},
  journal={arXiv preprint arXiv:2203.11854},
  year={2022}
}

@inproceedings{sionnart,
  title={Sionna {RT}: Differentiable ray tracing for radio propagation modeling},
  author={Hoydis, Jakob and A{\"\i}t Aoudia, Fay{\c{c}}al and Cammerer, Sebastian and Nimier-David, Merlin and Binder, Nikolaus and Marcus, Guillermo and Keller, Alexander},
  booktitle={IEEE Globecom Workshops (GC Wkshps)},
  year={2023},
  volume={},
  number={},
  pages={317-321},
}

@misc{openstreetmap,
  title={{OpenStreetMap}},
  url={https://www.openstreetmap.org/},
  note={Accessed: 2026}
}
\bibliographystyle{sciencemag}

%
%
%
%
%
%


\section*{Acknowledgments}
\paragraph*{Funding:}
This work was supported in part by Natural Science Foundation of China under Grants 62394292 and 624B2129, Ministry of Industry and Information Technology under Grant TC220H07E, and the Fundamental Research Funds for the Central Universities No. 226-2024-00069.
\paragraph*{Author contributions:}
Z.C. conceived the core idea and designed the methodology. Z.C. and H.R. implemented the technical framework, conducted the experimental evaluation, and wrote the main manuscript. Z.Z. inspired and supported the study, originated the basic idea and technical concepts, guided the experimental implementation and analysis, and contributed to the manuscript writing. H.R., Z.X., and R.L. were responsible for the data generation and visualization of figures. Z.Y. reviewed and revised the manuscript for technical accuracy and clarity. M.D. provided support for the experimental implementation. All authors discussed the results and contributed to the final version of the manuscript.
\paragraph*{Competing interests:}
The authors declare that there is no competing interest related to this paper.
\paragraph*{Data and materials availability:}
The code and data supporting the claims in this paper are available at \url{https://github.com/ziruichen-research/ALLoc}. All other data needed to evaluate and reproduce the results in the paper are present in the paper and/or the Supplementary Materials.


\subsection*{Supplementary materials}
Supplementary Text\\
Figs. S1 to S5\\
Tables S1 to S2\\
References \textit{(38-\arabic{enumiv})}\\ 


\newpage


\renewcommand{\thefigure}{S\arabic{figure}}
\renewcommand{\thetable}{S\arabic{table}}
\renewcommand{\theequation}{S\arabic{equation}}
\renewcommand{\thepage}{S\arabic{page}}
\setcounter{figure}{0}
\setcounter{table}{0}
\setcounter{equation}{0}
\setcounter{page}{1} 


\begin{center}
\section*{Supplementary Materials for\\ \scititle}

Zirui~Chen$^{1,2\dagger}$,
Hongning~Ruan$^{1,2,3\dagger}$,
Zhaoyang~Zhang$^{1,2,3\ast}$,
Ziqing~Xing$^{1,2,3}$,\\
Ridong~Li$^{1,2}$,
Zhaohui~Yang$^{1,2}$,
Mérouane~Debbah$^{4}$\\
\small$^{1}$College of Information Science and Electronic Engineering, Zhejiang University, Hangzhou 310027, China.\\
\small$^{2}$Zhejiang Provincial Key Laboratory of Multi-modal Communication Networks\\ and Intelligent Information Processing, Hangzhou 310027, China.\\
\small$^{3}$Institute of Fundamental and Transdisciplinary Research, Zhejiang University, Hangzhou 310058, China.\\
\small$^{4}$Research Institute for Digital Future, Khalifa University, 127788 Abu Dhabi, UAE.\\
\small$^\ast$Corresponding author. Email: {ning\_ming@zju.edu.cn}\\
\small$^\dagger$These authors contributed equally to this work.
\end{center}

\subsubsection*{This PDF file includes:}
Supplementary Text\\
Figures S1 to S5\\
Tables S1 to S2

\newpage



\subsection*{Supplementary Text}

\subsubsection*{Backgrounds and related works of intelligent wireless localization}
Wireless localization refers to inferring user location based on the user's wireless fingerprint, such as CSI between user and BS. Compared to GNSS, wireless localization offers inherent advantages in both precision and real-time performance, owing to shorter signal transmission distances and the inclusion of phase information \cite{cooper_loc}. 

Since the CSI typically is a complex coupling of multipath channel responses \cite{channel_deduction}, the location-related features it contains, such as amplitude attenuation and transmission angles, are often difficult to explicitly extract. Moreover, the unknown scattering environment makes it highly challenging to infer spatial location from these features using traditional signal processing algorithms. Recently, the accuracy of wireless localization is improving significantly empowered by AI \cite{wireless_loc_survey}. Using neural networks to learn a mapping function from wireless fingerprint to location, intelligent localization has allowed centimeter-level accuracy based on high-density measurement data \cite{ad_cnn1}. Moreover, researchers further explored numerous methods to optimize the accuracy \cite{ad_cnn2,ad_cnn2_long,ad_cnn3_long,cnn4,mccnet}, improve system compatibility \cite{fdma_positioning}, and reduce the cost of measurement \cite{transloc1,transloc2,metaloc}.

To address cross-scenario performance degradation in wireless AI, both initialization-based and feature-based cross-scenario model reuse methods have been studied. Leveraging transfer learning \cite{transfer_learning} or meta-learning \cite{meta_learning} techniques, some works improve the parameter initialization of LNN training to accelerate the learning process in new scenarios \cite{transloc1,transloc2,meta_learning}. Meanwhile, multi-task learning \cite{multi_task} has also been introduced into wireless AI research by treating different scenarios as different tasks \cite{multi_task_csinet}, which is also a potential means to optimize intelligent localization. By splitting the LNN into backbone and head, the backbone part is shared across scenarios and the head changes with the scenarios, so as to reduce the learning complexity in new scenarios. However, these methods still require training in new scenarios to tune the learned intelligence, which mitigates but cannot fundamentally resolve issues like training hardware cost and latency.

\subsubsection*{Differences between cross-scenario generalization and related learning paradigms}
Domain adaptation aims to maximize the model performance on a given target domain $\mathbf{D}_{\rm tar}$ using a training source domain $\mathbf{D}_{\rm src}$, where the data distributions of these two domains are different, i.e., $P_{\rm src}\ne P_{\rm tar}$. In domain generalization, the model is trained on $k$ source domains $\mathbf{D}_1,\dots,\mathbf{D}_k$ with distinct data distributions, i.e., $P_i\ne P_j~(i\ne j)$, and then generalizes to an unseen test domain $\mathbf{D}_{k+1}$ with the data distribution $P_{k+1}\ne P_i$ (for $i\in\{1,\dots,k\}$). The difference between domain adaptation and domain generalization is that domain adaptation allows for additional training or tuning using data from the target domain, whereas domain generalization does not. However, in both problems, the mapping relationship $\mathrm{g}(\cdot)$ remains invariant across domains. This makes domain adaptation or domain generalization problems more tractable, because the distribution differences do not hinder joint learning of the shared $\mathrm{g}(\cdot)$ from multiple domains, and the feasibility of joint learning also facilitates generalization to new distributions. In contrast, in cross-scenario generalization problem studied in this article, the mapping function $\mathrm{g}(\cdot)$ varies across scenarios, making it impossible to fully share the learning process between them. Even if $\mathrm{g}_{S_1}(\cdot),\dots,\mathrm{g}_{S_k}(\cdot)$ have been fitted by neural networks, these complex black-box results cannot form effective ensemble akin to data distributions to meet the functional requirements of $\mathrm{g}_{S_{k+1}}(\cdot)$.

Transfer learning involves training a model on a source task and then improving its performance on a different but related target task. Model fine-tuning is a common strategy in transfer learning. Multi-task learning, on the other hand, involves jointly optimizing a model across multiple related tasks. By sharing representations across these tasks, the model achieves better generalization and facilitates adaptation to new tasks. Both transfer learning and multi-task learning can handle tasks with different mapping functions $\mathrm{g}(\cdot)$, so we use them as comparative methods for multi-scenario wireless localization in our experiments. However, both methods require additional training or fine-tuning on the target task. In multi-scenario problems, due to the large number of scenarios and their dynamic nature, conducting additional training for each scenario would incur significant computational cost and latency. Therefore, cross-scenario generalization is a critical and highly challenging problem.

\subsubsection*{Data-to-label baselines}
The comparison methods in this work include MFCNet \cite{fdma_positioning}, AD\_CNN \cite{ad_cnn2_long}, and CNN \cite{cnn4}. Additionally, transfer learning \cite{transfer_learning} and multi-task learning \cite{multi_task} are used as comparison methods to evaluate the generalization capability of AL. These methods can be outlined as follows:
\begin{itemize}
    \item MFCNet: This method employs LSTM as the backbone network, leveraging information accumulation across multiple carriers to map MIMO-OFDM CSI to user location. It is a method that has demonstrated state-of-the-art accuracy.
    \item AD\_CNN: This method first transforms CSI matrix into sparse angle-delay domain using two-dimensional discrete Fourier transform (2D DFT), and then maps it to user location using a CNN. It is one of the most widely-used LLNs \cite{ad_cnn1,ad_cnn2_long,ad_cnn3_long}.
    \item CNN: This method omits the DFT step in AD\_CNN and directly maps CSI to user location using a CNN. It is also a widely used LNN \cite{cnn1,cnn2,cnn4}.
    \item Transfer learning: This method involves using the parameters trained in another scenario as the initial parameters of the LNN in a new scenario, and then fine-tuning the model with new scenario's data to accelerate the learning process. 
    \item Multi-task learning: This method splits the LNN into a shared backbone network and scenario-specific heads. The backbone network shares parameters across all scenarios, while each scenario's head is trained independently. During pre-training, multi-scenario data is used to jointly train the backbone network. When deploying to a new scenario, a new head is trained using local data.
\end{itemize}

MFCNet, AD\_CNN, and CNN differ only in their network architectures, which results in slight variations in final accuracy. To clearly presenting results, we set the widely-used AD\_CNN as the default LNN's network architecture in transfer learning and multi-task learning.

\subsubsection*{Comparisons between AL and ICL}
ICL refers to the AI models to perform specific tasks by leveraging contextual information provided in the input, such as task descriptions or examples.
ICL has demonstrated remarkable utility in LLMs. However, in non-language fields, where tasks are often difficult to precisely characterize, sample-based contextual prompts are typically employed.  Recently, researchers have preliminarily explored the feasibility of ICL in linear or approximately linear wireless problems, including analyzing its performance in estimators and equalizers \cite{in_context_estimator,in_context_equalization}. Nevertheless, current researches still lack attempts to apply ICL to highly-nonlinear problems such as wireless localization, as well as evaluation of its results in multi-scenario case.

\begin{figure} 
	\centering
	\includegraphics[width=\textwidth]{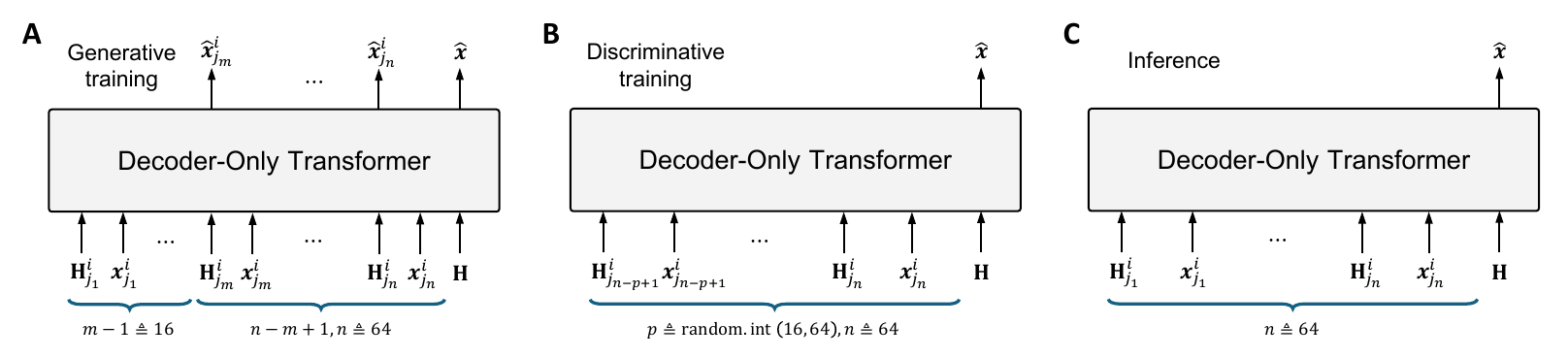}
	\caption{\textbf{The schematic diagram of applying in-context learning to wireless localization.}
		(\textbf{A} and \textbf{B}) Two different training schemes, treating ICLLoc as a sequence generation task (A) and a label discrimination task (B). (\textbf{C}) The final inference process.}
	\label{fig:in_context}
\end{figure}

This work applies the typical paradigm of ICL to wireless localization, as illustrated in Fig. \ref{fig:in_context}. The model employs a decoder-only Transformer, learnable positional embedding, and causal masking. To ensure fair comparisons, the model size and depth of ICLLoc's Transformer are kept consistent with those of Mateformer, while the hidden size is doubled, as Mateformer incorporates two Transformers. Training of ICLLoc is conducted under the same data and parameter settings as ALLoc. In each training step, the paired sets $\{\mathbf{H}\}$ and $\{\bm{x}\}$ (each containing $n+1$ elements) retrieved from the training set are shuffled internally, while the order correspondence between the sets is preserved. This diversifies the combinations of contextual information and aligns with the two-part random sampling in ALLoc's training. To ensure consistency between training and inference, the sample to be inferred is always assigned to the last positional embedding. 

In fact, one perspective on why ICL can derive effective prompts from shots is that the model possesses a certain degree of analogical reasoning capability \cite{in_context_survey}. The effectiveness of AL further illustrates the soundness of this perspective. In Mateformer, the output never receives absolute features from $\mathbf{H}$ but only relative weights. This forces the model to rely on analogical reasoning rather than reverting  into a data-to-label approach. Meanwhile, the AL approach significantly enhances the model's adherence to analogical mechanisms through built-in architectural priors. Specifically, compared to Transformer, Mateformer offers at least two advantages in learning Eq. \ref{eq:new_paradigm}.
\begin{itemize}
    \item \textbf{Strict data-label correspondence}: Mateformer divides $\{\mathbf{H}_{j_1}^i,\bm{x}_{j_1}^i,\dots,\mathbf{H}_{j_n}^i,\bm{x}_{j_n}^i,\mathbf{H}\}$ into two sequences,  $\{\mathbf{H}_{j_1}^i,\dots,\mathbf{H}_{j_n}^i,\mathbf{H}\}$ and  $\{\bm{x}_{j_1}^i,\dots,\bm{x}_{j_n}^i\}$, and defines the relativity weights of $\mathbf{H}_j^i$ for $\bm{x}_j^i$'s transformation, thereby structurally  characterizing the correspondence between $\mathbf{H}_j^i$ and  $\bm{x}_j^i$. In contrast, Transformer-based ICL methods lack such pre-definitions and can only rely on positional embeddings to indicate correspondences. This not only compromises precision but also increases the learning burden.
    \item \textbf{Focused attention mechanism}: Mateformer includes two separate Transformers, each with a defined purpose to focus on $\{\mathbf{H}_{j_1}^i,\dots,\mathbf{H}_{j_n}^i,\mathbf{H}\}$ and  $\{\bm{x}_{j_1}^i,\dots,\bm{x}_{j_n}^i\}$ respectively. Differently, when directly applying a Transformer to learn the whole sequence $\{\mathbf{H}_{j_1}^i,\bm{x}_{j_1}^i,\dots,\mathbf{H}_{j_n}^i,\bm{x}_{j_n}^i,\mathbf{H}\}$, the model may still attend to irrelevant data-label combinations, such as $\mathbf{H}_j^i$\&$\bm{x}_{j'}^i$ (where $j>j'$), which lack meaningful correspondence. This significantly reduces the efficiency of attention extraction and synthesis.
\end{itemize}
These properties and the resulting performance improvements are significant for the further development of contextualized sample embedding techniques. 
In addition, mainstream ICL is more often defined as a phenomenon rather than a learning paradigm; it is regarded as an emergent capability of LLMs acquired after large-scale pretraining. In contrast, the proposed ALLoc method and the introduced ICLLoc baseline maintain alignment between training and deployment stages, both explicitly constructing contextual information. Benefiting from the guidance of prior structural knowledge, the AL approach requires only a small number of samples—even with single cell data—and still achieves strong performance and good generalization.

\subsubsection*{Robustness evaluation and analysis}
\begin{figure} 
	\centering
	\includegraphics[width=\textwidth]{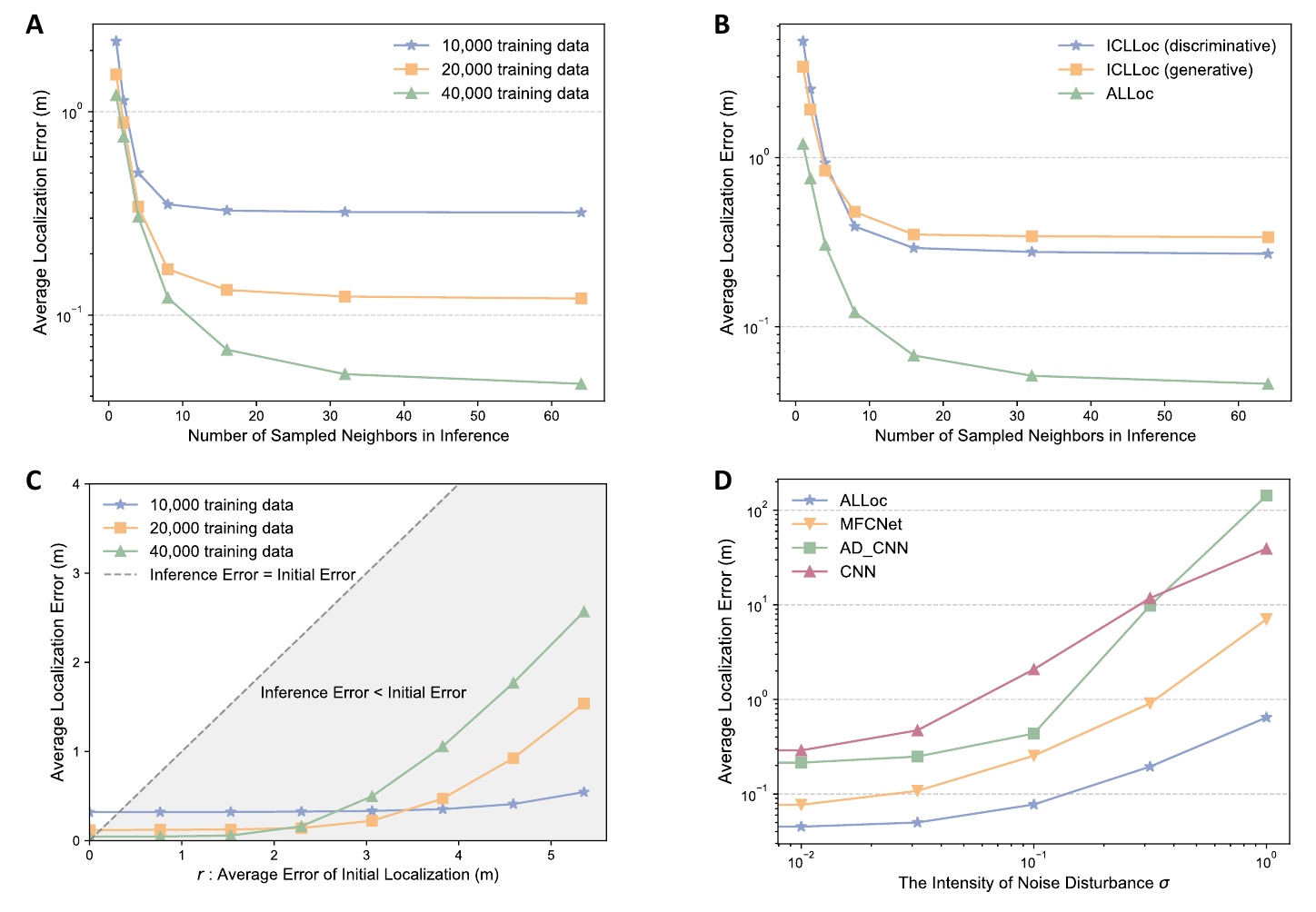}
	\caption{\textbf{Robustness evaluation in `O1' scenario.}
		(\textbf{A} and \textbf{B}) Localization accuracy versus number of sampled neighbors. (\textbf{C}) ALLoc's localization accuracy versus accuracy of initial location error, where $r$ is calculated as in Eq. \ref{eq:init_loc_error}. (\textbf{D}) Evaluation of LNNs' robustness to noisy input. }
	\label{fig:results_analysis}
\end{figure}

We conduct several experiments in `O1' scenario to evaluate the robustness of ALLoc. Figs. \ref{fig:results_analysis}A and \ref{fig:results_analysis}B shows the relationship between localization accuracy and the number of neighborhood samples: the localization error decreases initially and then saturates as the number of neighborhood samples increases, indicating that sufficient neighborhood information is conducive to accurate location synthesis. Meanwhile, ALLoc demonstrates a clear performance advantage over ICLLoc in multi-aspects, including both accuracy and the efficiency of utilizing embedded information.

Fig. \ref{fig:results_analysis}C illustrates the relationship between ALLoc's localization accuracy and the initial coarse location accuracy. As the initial error increases, the final localization performance first remains stable without significant decline and then gradually degrades. Moreover, a distinct phenomenon is observed: under larger initial errors, the ALLoc's localization performance with fewer training samples outperforms that with more, demonstrating a reversal in performance superiority. These phenomena are essentially derived from the degree of overlap between the true neighborhood and the pseudo neighborhood. When the initial error is small, the true neighborhood and the pseudo neighborhood always have a large overlap, which allows ALLoc to still obtain effective neighborhood scenario embeddings. On the contrary, when the initial error is too large, the pseudo neighborhood can no longer approximate the true neighborhood. In this case, the larger the initial error, the less effective the embedded neighborhood, leading to an increase in the localization error of ALLoc. Additionally, the fewer training samples, the larger average spatial intervals between each known sample. Under the same number of neighborhood samples, the coverage area becomes wider, which in turn increases the similarity between the pseudo neighborhood and the true neighborhood, thereby achieving better performance.

Fig. \ref{fig:results_analysis}D demonstrates the localization accuracy when the input CSI are noise-contaminated rather than ideal. The disturbed CSI $\mathbf{H}_{\text{dis}}$ is obtained by applying the Hadamard product ($\odot$) between the ideal CSI $\mathbf{H}$ and a Gaussian random noise matrix $\mathbf{D}$, where each element of $\mathbf{D}$ follows $N(1, \sigma^2)$. Here, $\sigma$ controls the noise intensity, representing the deviation between the noisy and ideal channels \cite{fdma_positioning}. ALLoc demonstrates superior noise resistance compared to baseline methods, which is highly valuable for ensuring reliable performance in practical applications.

\newpage


\begin{figure} 
	\centering
	\includegraphics[width=\textwidth]{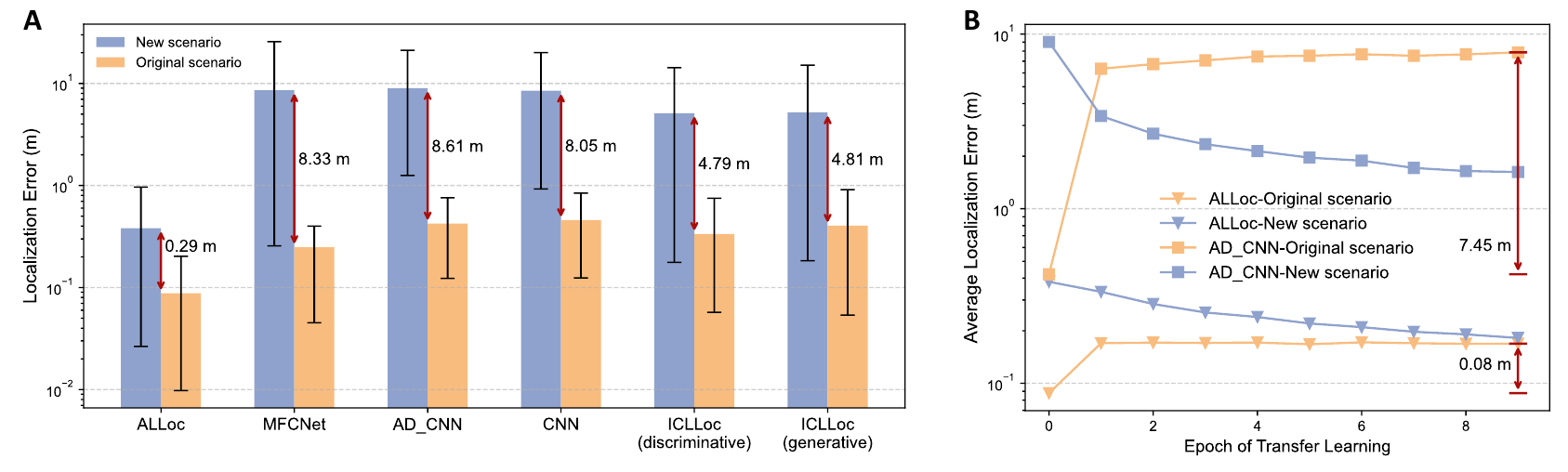}
	\caption{\textbf{Supplementary experimental results regarding cross-scenario generalization.}
		(\textbf{A}) Direct reusing the model trained in `O1B' scenario to `O1' scenario. (\textbf{B}) Transfer learning with the model parameters trained in `O1B' scenario as initialization and the scenario data of `O1' scenario as training set.}
	\label{fig:results_deepmimo_o1b_to_o1}
\end{figure}

\begin{figure} 
	\centering
	\includegraphics[width=\textwidth]{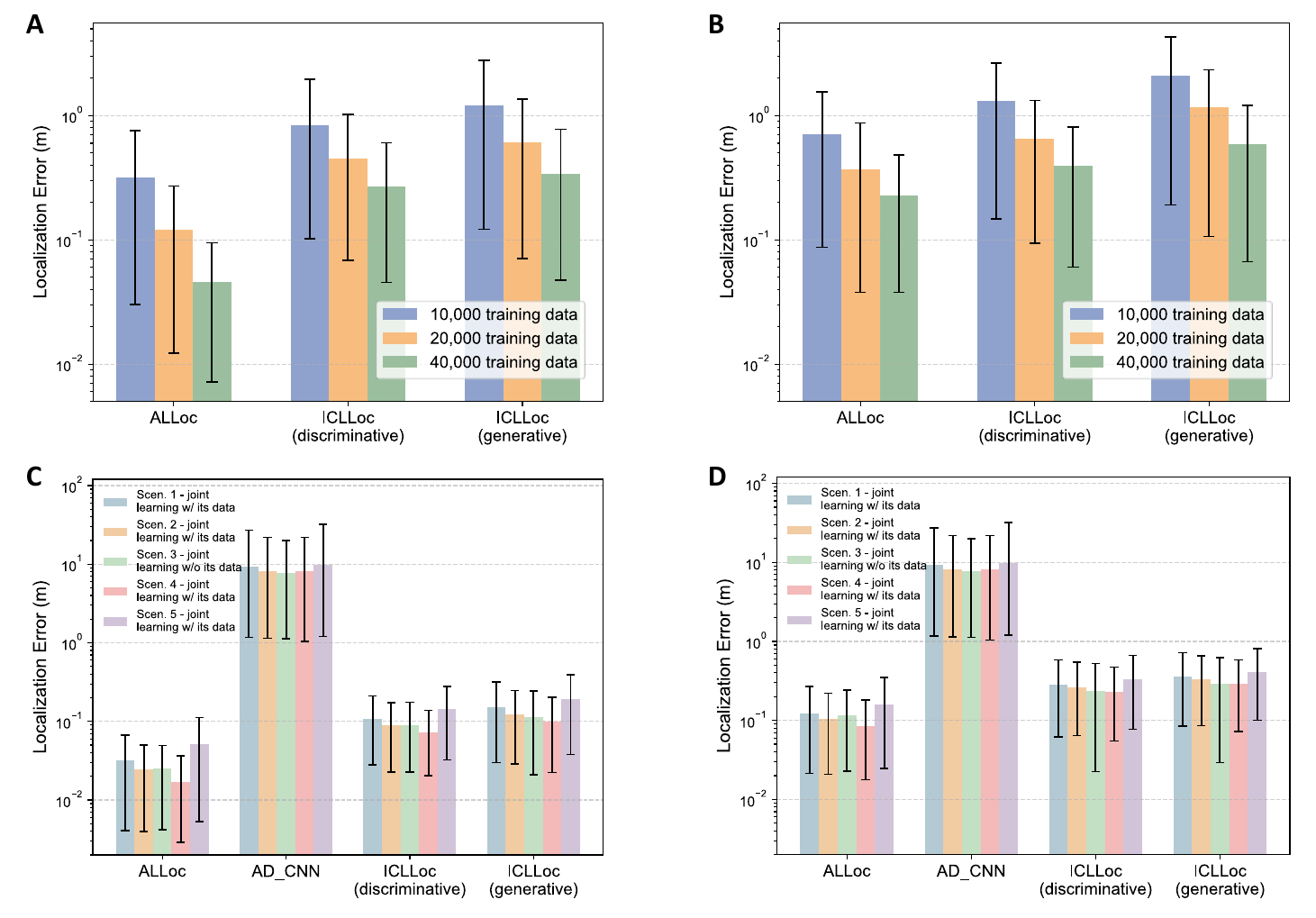}
	\caption{\textbf{Localization accuracy under neighborhood sampling or random sampling.}
		(\textbf{A} and \textbf{B}) Localization error in `O1' scenario based on neighborhood sampling (A) and random sampling (B). (\textbf{C} and \textbf{D}) Localization error in `MO1' scenario based on neighborhood sampling (C) and random sampling (D).}
	\label{fig:results_neighbor_random}
\end{figure}

\begin{figure} 
	\centering
	\includegraphics[width=\textwidth]{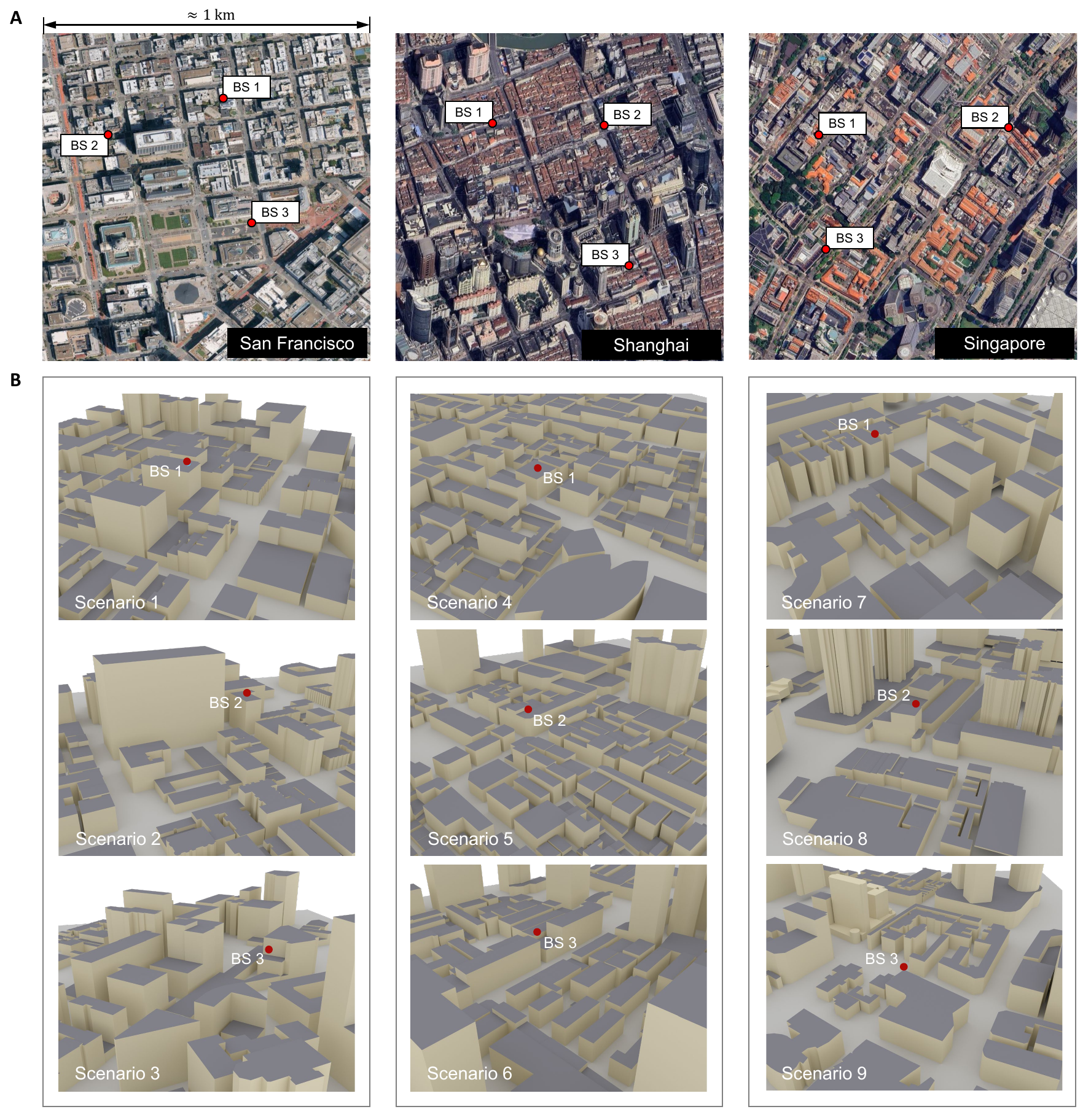}
	\caption{\textbf{Realistic urban scenarios used in experiments.}
		(\textbf{A}) Satellite images of the three cities, namely San Francisco, Shanghai, and Singapore. (\textbf{B}) Overview of the 9 scenarios. Scenarios 1 to 3 belongs to San Francisco, scenarios 4 to 6 belongs to Shanghai, and scenarios 7 to 9 belongs to Singapore. The 3D geographic data is retrieved from OpenStreetMap \cite{openstreetmap}.}
	\label{fig:map}
\end{figure}


\begin{table} 
	\centering
	\caption{\textbf{Parameter information of used DeepMIMO datasets.}}
	\label{tab:deepmimo}
	\begin{tabular}{lllll}
\hline
Scenario & BS & User area & {Number of training data} & {Number of testing data} \\
\hline
`O1' and `O1B' & BS3 & R501-R1400 & {10,000/20,000/40,000 } & 20,000 \\ 
 & & & (default: 40,000) & \\ 
`MO1' scen.1 & BS2 & R101-R700 & 20,000 & 20,000 \\ 
`MO1' scen.2 & BS3 & R701-R1200 & 20,000 & 20,000 \\ 
`MO1' scen.3 & BS5 & R1201-R1600 & 20,000 & 20,000 \\ 
`MO1' scen.4 & BS8 & R1601-R2000 & 20,000 & 20,000 \\ 
`MO1' scen.5 & BS9 & R2001-R2700 & 20,000 & 20,000 \\ 
`RO1' & BS3 & R501-R1400 & 40,000 & 20,000 \\ 
`O2' (each cond.) & BS1 & R1-R31 & 20,000 & 20,000 \\ 
\hline
	\end{tabular}
\end{table}

\begin{table} 
	\centering
	\caption{\textbf{Parameter Settings of ALLoc in the Experiments.}}
	\label{tab:model}
	\begin{tabular}{p{6cm}p{7cm}}
\hline
\textbf{Parameters} & \textbf{Value} \\
\hline
Batch size & 500 \\
Training steps & 300,000 \\
Learning rate & {1 $\times 10^{-4}$ (initial), multiplied by 0.2 every 50,000 steps after the 100,000th step.} \\
Loss function & MSE \\	
Optimizer & Adam \cite{adam}   \\ 
Number of total neighbors ($n$) & 64 \\
Number of embedding samples in training ($p$) & Randomly chosen in [16,64] \\
Number of to-be-inferred samples in training ($q$) & Randomly chosen in [1,64] \\
\hline
Hyperparameters of Mateformer & ~ \\
Depth ($N$) & 8 \\
Model size & 256 ($d_{\rm model}$ of Transformer) \\
Hidden size & 256 ($d_{\rm ff}$ of Transformer) \\
Number of heads & 4 ($h$ of Transfomer) \\
Pre\_norm or Post\_norm & Pre\_norm \cite{pre_norm} \\
\hline
	\end{tabular}
\end{table}


\clearpage 

\end{document}